%% file: root.tex
\documentclass[letterpaper, 10 pt, conference]{ieeeconf}  

\IEEEoverridecommandlockouts                              

\usepackage{graphics} 
\usepackage{epsfig} 
\usepackage{mathptmx} 
\usepackage{times} 
\usepackage{amsmath} 
\usepackage{amssymb}  

\newcommand{\etal}{\textit{et al.}}
\usepackage{graphicx} 
\usepackage{subcaption}
\usepackage{multirow}
\usepackage{amssymb}
\usepackage{hyperref}
\usepackage{caption}
\usepackage{float}

\title{\LARGE \bf
MEDL-U: Uncertainty-aware 3D Automatic Annotation based on Evidential Deep Learning
}

\author{Helbert Paat, Qing Lian, Weilong Yao and Tong Zhang
\thanks{Helbert Paat, Qing Lian, and Tong Zhang are with the Department of Computer Science and Engineering, Hong Kong University of Science and Technology, Hong Kong, China. {\tt\small hpaat, qlianab}@connect.ust.hk, {\tt\small tongzhang}@ust.hk. Weilong Yao is with Autowise.AI. {\tt\small yaoweilong}@autowise.ai}
}

\begin{document}

\maketitle
\thispagestyle{empty}
\pagestyle{empty}
\vspace*{-\baselineskip}

\begin{abstract}

Advancements in deep learning-based 3D object detection necessitate the availability of large-scale datasets. However, this requirement introduces the challenge of manual annotation, which is often both burdensome and time-consuming. To tackle this issue, the literature has seen the emergence of several weakly supervised frameworks for 3D object detection which can automatically generate pseudo labels for unlabeled data. Nevertheless, these generated pseudo labels contain noise and are not as accurate as those labeled by humans. In this paper, we present the first approach that addresses the inherent ambiguities present in pseudo labels by introducing an Evidential Deep Learning (EDL) based uncertainty estimation framework. Specifically, we propose MEDL-U, an EDL framework based on MTrans, which not only generates pseudo labels but also quantifies the associated uncertainties. However, applying EDL to 3D object detection presents three primary challenges: (1) relatively lower pseudo label quality in comparison to other autolabelers; (2) excessively high evidential uncertainty estimates; and (3) lack of clear interpretability and effective utilization of uncertainties for downstream tasks. We tackle these issues through the introduction of an uncertainty-aware IoU-based loss, an evidence-aware multi-task loss, and the implementation of a post-processing stage for uncertainty refinement. Our experimental results demonstrate that probabilistic detectors trained using the outputs of MEDL-U surpass deterministic detectors trained using outputs from previous 3D annotators on the KITTI val set for all difficulty levels. Moreover, MEDL-U achieves state-of-the-art results on the KITTI official \textit{test} set compared to existing 3D automatic annotators.


\end{abstract}

\section{INTRODUCTION}
\input{p1.intro.tex}
\section{Related Literature}

\input{p2.rl.tex}
\input{p3.method.tex}
\input{p4.exp.tex}
\input{p5.con.tex}

\bibliographystyle{IEEEtran}
\bibliography{IEEEabrv,root}

\clearpage
\section*{Appendix}
\renewcommand{\thesubsection}{\Alph{subsection}}
\renewcommand\thefigure{\thesubsection.\arabic{figure}}    

\input{appendix.tex}










\end{document}

%% file: p1.intro.tex
Localizing 3D objects in world coordinates is a fundamental module in many robotics and autonomous driving applications. Recently, with the development of deep neural networks, network-based methods \cite{Qian20213DOD} have dominated this field and are capable of classifying, detecting, and reconstructing objects in 3D space.

\begin{figure}[!t]
    \begin{center}
        \includegraphics[width=9cm]{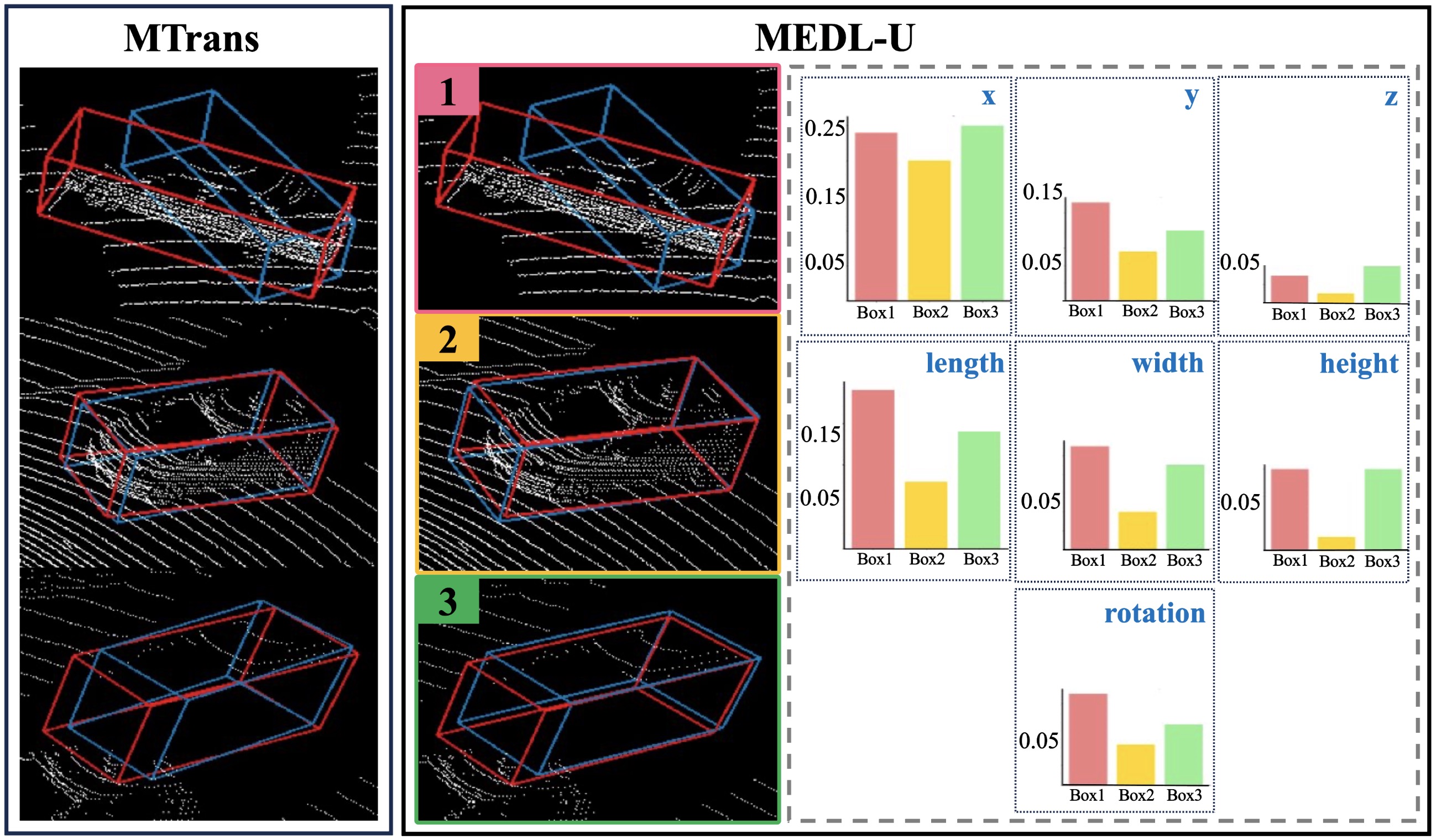} 
        \caption{Illustration of the proposed MEDL-U in comparison with current state-of-the-art 3D autolabeler, MTrans \cite{Liu2022MultimodalTF}. MEDL-U not only generates pseudo labels but also estimates the associated uncertainties to indicate the inaccuracy of the pseudo labels. Ground-truth boxes and pseudo labels are colored red and blue, respectively.}
        \label{motiv_problem}
        \end{center}
        \vspace*{-\baselineskip}
        \vspace*{-\baselineskip}
\end{figure}

However, the training of network-based 3D detectors requires a massive amount of data labeled with 3D bounding boxes, which often involves significant costs \cite{Song2015SUNRA, Wang2018TheAO}. To alleviate the heavy annotation burden, one promising direction is weakly supervised training that utilizes LiDAR data, image, and 2D bounding boxes to train a 3D object annotator \cite{Wei2021FGRFG, Liu2022MAPGenAA, Liu2022MultimodalTF, Qian2023ContextAwareTF}. The weakly-supervised methods propose frameworks that can automatically annotate objects in 3D, minimizing the reliance on ground truth labels during downstream training of 3D detectors. Although current approaches can achieve good 3D bounding box annotations, the generated 3D bounding boxes are not as accurate as those labeled by humans. In the illustrated pseudo labels from MTrans \cite{Liu2022MultimodalTF} on the left side of Figure \ref{motiv_problem}, it is evident that pseudo labels 1 and 3 contain imprecise estimates of box parameters. Unfortunately, current approaches neglect these annotation noises, directly utilizing these pseudo labels to train 3D detectors. Clearly, neglecting these noises in pseudo labels degrades the effectiveness of training downstream 3D detectors. To alleviate this problem, our work considers both the tasks of annotating the 3D bounding boxes and estimating the annotation uncertainty to indicate the annotation inaccuracies. On the right side of Figure \ref{motiv_problem}, we show that our work not only predicts pseudo labels but also determines the uncertainty estimates for each 3D box parameter, which are then utilized to train 3D detectors more effectively.




Evidential deep learning (EDL) has been effectively utilized for uncertainty estimation in regression tasks \cite{Amini2019DeepER} and has found diverse applications in computer vision tasks \cite{Wang2021UncertaintyEF, Bao2021EvidentialDL, Gawlikowski2022AnAD}. Hence, we propose \textbf{M}Trans-based \textbf{E}vidential \textbf{D}eep \textbf{L}earning autolabeler with \textbf{U}ncertainty Estimation capability (MEDL-U). Our input is similar to the typical 3D automatic annotator, which comprises of a collection of scene frames, corresponding LiDAR data, 2D image, and the 2D bounding boxes for the objects. With these inputs, our goal is to develop a model that produces not only accurate 3D bounding box annotations for the surrounding objects but also a measure of uncertainty for annotated bounding box parameters (the predicted center, length, width, height, and rotation (yaw angle)), all the while avoiding any additional manual annotation costs or huge computational overhead. 


However, directly applying the EDL framework for uncertainty estimation in 3D autolabeler introduces three main challenges: (1) directly incorporating the evidential loss with evidence regularizer from EDL framework \cite{Amini2019DeepER} results in worse performance during inference compared to the IoU-based loss as the latter unifies all 3D box parameters into one metric and aligns with the evaluation objective; (2) the uncertainties are not well-calibrated and can become unreasonably high during training; and (3) lack of interpretability, reasonable applicability and proper utilization of the generated uncertainties due to the variations in the magnitude of the evidential parameters for each 3D box.

To address these problems, we introduce an uncertainty-aware IoU loss to help the model regress high-quality box variables. Moreover, we make the multi-task loss functions evidence-regularized with the intuition that the model's predicted total evidence, as determined by the EDL framework, is inversely related to the losses for multiple tasks. Finally, we propose a post-processing step involving the rescaling of uncertainties to ensure uniformity across diverse box parameters. Simultaneously, we incorporate an objective centered on minimizing a monotonic function, denoted as f and parameterized by $\kappa$, whose primary aim is to determine the optimal value of $\kappa$ that, upon applying the uncertainties through f, yields the lowest Negative Log-Likelihood (NLL) over the same limited training dataset utilized to train the 3D autolabeler.

With a limited number of annotated frames (e.g. 500 frames), our proposed MEDL-U not only generates 3D box annotations but also measures of uncertainty  for each pseudo label box parameter, which can be utilized for loss reweighting during the training of existing 3D object detectors. Extensive experiments demonstrate that our MEDL-U improves the performance of 3D detectors during inference on the KITTI \textit{val} and \textit{test} set and outperforms previous 3D autolabelers. 

%% file: p2.rl.tex
\subsection{Automatic 3D Bounding Boxes Annotation}
\label{}

Recently, the literature has witnessed a rise in 3D automatic annotation frameworks. An example is WSPCD \cite{Meng2021TowardsAW} which allows learning 3D object parameters from a few weakly annotated examples. It has a two-stage architecture design: the first stage for cylindrical object proposal generation and the second stage for cuboids and confidence score prediction. A non-learning based approach that detects vehicles in point clouds without any 3D annotations is FGR \cite{Wei2021FGRFG} which also consists of two stages: coarse 3D segmentation stage and the bounding box estimation stage. More recently, Liu \etal \cite{Liu2022MultimodalTF} propose a Transformer-based 3D annotator called MTrans, which aims to address the prevalent sparsity problem of unstructured point clouds from LiDAR scans by generating extra 3D points through a multimodal self-attention mechanism combined with additional multi-task and self-supervision design. Different from previous appoaches, Qian \etal \cite{Qian2023ContextAwareTF} propose a simplified end-to-end Transformer model, CAT, which captures local and global relationships through an encoder-decoder architecture. The encoder consists of intra-object encoder (local) and inter-object encoder (global) which performs self-attention along the sequence and batch dimensions. Through this, it can model the inter-object feature relation that gives additional information to hard samples. Additionally, several approaches (GAL \cite{Yin2023GALGA}, VS3D \cite{Qin2020WeaklyS3}, WS3DPR \cite{Liu2022EliminatingSA}) have also been proposed. However, all these 3D automatic annotators only generate 3D pseudo labels without any estimation of the uncertainty or noise associated with them. In this study, we utilize the generatd 3D pseudo labels and the estimated uncertainties to train 3D detectors. This approach aims to mitigate the impact of generated noisy labels by reducing the influence of inaccurate supervision signals and enabling the model to effectively learn from more reliable pseudo labels.

\subsection{Uncertainty Estimation and 3D Probabilistic Detection}

Uncertainties in deep learning-based predictions can be categorized into two: one that is caused by inherent noise in data (aleatoric), and the other is model uncertainty due to incomplete training or model design (epistemic). As a tool for uncertainty estimation in regression tasks \cite{Amini2019DeepER}, Evidential Deep Learning (EDL) has found diverse applications in various tasks such as stereo matching \cite{Wang2021UncertaintyEF}, open set recognition \cite{Bao2021EvidentialDL}, molecular structure prediction \cite{Soleimany2021EvidentialDL}, and remote sensing \cite{Gawlikowski2022AnAD}. In this work, we use the framework of EDL to estimate prediction uncertainties in 3D Object Detection. Utilizing these uncertainties to define pseudo label distributions, probabilistic object detectors, which typically adapt deterministic detectors' architecture, can enable the prediction of probability distributions for object categories and bounding boxes. A framework that utilizes probabilistic detectors is GLENet \cite{Zhang2022GLENetB3}, where the 3D detector predict distributions for the 3D box and assume that the ground truth labels follow Gaussian distribution with the uncertainties as the variance. They train the models with the KL divergence loss to supervise the predicted localization uncertainty. In this work, we follow the same approach but instead incorporate 3D pseudo label uncertainties.

%% file: p3.method.tex
\section{Evidential Deep Learning (EDL) for Uncertainty Estimation in 3D Automatic Labelers}
\label{edl_method}

\subsection{Automatic Annotation with Pseudo label Uncertainty Estimation}
\label{edl_auto_anno}

Given the point cloud data, 2D image and the 2D bounding boxes of each object, the objective of this work is to generate the 3D bounding boxes annotation for each object and estimate the corresponding uncertainty for each 3D box parameter. First, the autolabeler is initially trained with a small set of ground truth 3D bounding boxes (e.g., 500 frames of data), where the input of the autolabeler are the point cloud data, 2D image and the 2D bounding boxes of each object and the output are the estimated 3D bounding boxes and corresponding uncertainty estimates. Secondly, with the trained autolabeler, we employ it to predict the 3D bounding boxes for the remaining data and do the uncertainty estimation for the predicted 3D boxes. Finally, we leverage the predicted 3D bounding boxes and estimated uncertainty to train a downstream probabilistic 3D detector on the massive weakly annotated data. Compared to fully supervised setting, our work only needs a few frames of labeled data to train the 3D autolabeler, which significantly reduces the manual annotation cost.

In this paper, we build the model architecture for the 3D automatic labeler from MTrans \cite{Liu2022MultimodalTF}. MTrans extracts object features using a multimodal self-attention module that processes point cloud and image inputs fused with point-level embedding vectors. The extracted object features are utilized for various tasks such as foreground segmentation, point generation, and 3D box regression. However, the generated 3D box pseudo labels may contain noise. To account for these inaccuracies of the pseudo labels, we incorporate uncertainty estimation task to MTrans by considering EDL, a powerful uncertainty estimation framework, and applying it to MTrans. For a straightforward incorporation of uncertainty estimation in MTrans via EDL, we include an evidential box head to regress the parameters of the evidential distribution and the 3D bounding box. For training the model, we replace the dIoU loss with the evidential loss to supervise the model in learning the box parameters. Moreover, the evidence regularizer is added to calibrate the uncertainties. However, there are problems with this approach of directly applying EDL in MTrans, which we will discuss in the next sections.



\subsection{Background on Evidential Deep Learning}
\label{edl_prob_setup}


In 3D object detection, we define a 3D bounding box by its center coordinates (\textit{x}, \textit{y} and \textit{z}), length (\textit{l}), width (\textit{w}), height (\textit{h}), and rotation (yaw angle denoted as $rot$). From the viewpoint of EDL, we assume each label $j \in \mathbb{J} = \{x,y,z,l,w,h,rot\}$ is drawn i.i.d. from a Gaussian distribution where the mean $\mu_j$ and variance $\sigma ^ 2_j$ are unknown. EDL framework assumes that $\mu_j$ is drawn from a Gaussian prior and $\sigma^2_j$ is drawn from an inverse-gamma prior.

\begin{center}
    $j \sim  \mathcal{N}(\mu_j, \sigma_j^2),  \:\:\:\:$
    $\mu_j \sim \mathcal{N}(\gamma_j, \sigma_j^2\nu_j^{-1}),\:\:\:\:$ $\sigma_j^2 \sim \Gamma^{-1}(\alpha_j, \beta_j)$
\end{center}

where $\gamma_j \in \mathbb{R}$, $\nu_j > 0$, and $\Gamma(\cdot)$ is the gamma function with $\alpha_j > 1$ and $\beta_j > 1$.


Let $\Phi_j$ and $\Theta_j$ denote the set of parameters $\{\mu_j, \sigma_j^2\}$ and $\{\gamma_j, \nu_j, \alpha_j, \beta_j\}$, respectively. Assuming independence of the mean and variance, the posterior $p(\Phi_j| \Theta_j)$ is defined to be an normal-inverse gamma (NIG) distribution, which is a Gaussian conjugate prior.



As presented in \cite{Amini2019DeepER}, the hyperparameters of the evidential distribution can be obtained by training a deep neural network (called evidential head) to output such values. For each 3D bounding box parameter, our model predicts four evidential parameters: $\gamma_j, \nu_j, \alpha_j, \beta_j$. 

Through an analytic computation of the maximum likelihood Gaussian without repeated inference for sampling \cite{Amini2019DeepER}, EDL provides a framework for the estimation of uncertainties in regression. We can calculate the prediction, aleatoric, and the epistemic uncertainties for each 3D box parameter as follows:
\vspace{-0.2cm} 

\begin{equation}
    E[\mu_j] = \gamma_j,  \:\:\:\:\:\:\:\:  E[\sigma_j ^2] = \dfrac{\beta_j}{\alpha_j - 1},
\end{equation}

\vspace*{-\baselineskip}
\begin{equation}
Var[\mu_j] = E[\sigma_j ^ 2] / v_j = \dfrac{\beta_j}{v_j (\alpha_j - 1)}.
\end{equation}
\vspace*{-\baselineskip}




\textit{Maximizing the data fit:}
We are given the hyperparameters $\Theta_j$ of the evidential distribution as outputs of the proposed evidential head for $j \in \mathbb{J} = \{x,y,z,l,w,h,rot\}$. The likelihood of an observation $y_j$ is computed by marginalising over the likelihood parameters $\Phi_j$. If we choose to impose a NIG prior onto the Gaussian likelihood function results, an analytical solution is derived as follows: 
\vspace{-0.3cm} 
\begin{equation}
\begin{aligned}
p(y_j|\Theta_j) = St_{2 \alpha_j} (y_j|\gamma_j, \dfrac{\beta_j (1+\nu_j)}{\nu_j \alpha_j}), 
\end{aligned}
\end{equation}
\vspace{-0.4cm} 

where $St_{\nu} (t|r,s)$  corresponds to the evaluation of the Student's t-distribution at the value t, with parameters for location ($r$), scale ($s$), and degrees of freedom ($\nu$). 



For training the EDL framework, we define the evidential loss as the mean of the negative log likelihood (NLL) for each 3D box parameter as follows:
\vspace{-0.2cm} 
\begin{equation}
\begin{aligned}
\mathcal{L}_{evi} (\Psi) & = -\dfrac{1}{|\mathbb{J}|} \sum\limits_{j \in \mathbb{J}} \text{ log } \: p(y_{j} | \Theta_j). \\
\end{aligned}
\end{equation}
\vspace{-0.4cm}




 \textit{Uncertainty Calibration:}
Similar to Amini \etal \cite{Amini2019DeepER}, we can scale the total evidence with the prediction error for each 3D box parameter in the following manner:
\vspace{-0.3cm} 

\begin{equation}
    \mathcal{L}_{R}(\theta) = \dfrac{1}{|\mathbb{J}|} \:  \sum\limits_{j \in \mathbb{J}} \phi_j \lVert y_j - \gamma_j \rVert ,
\end{equation}
\vspace{-0.3cm} 

where $\phi_j$ is the total evidence defined as $\phi_j = 2 \nu_j + \alpha_j$, and $\lVert \cdot \lVert$ is L1 norm.




\subsection{Problems with Utilizing EDL for 3D Bounding Box Regression} 
\label{edl_problems}


\begin{itemize}
    \item Significant variability exists in uncertainty estimates for box regression parameters, with some resulting in excessively high values. As also discussed in \cite{Oh2021ImprovingED}, there exist the gradient shrinkage problem in the evidential NLL loss, where the model can decrease the loss value by increasing the uncertainty instead of getting accurate point estimates for the predicted bounding box parameters. The gradient with respect to the the predicted values can become very small by merely increasing the uncertainties.    
      
    \item Since the NLL-based evidential loss is not sufficient to optimize the accuracy of the prediction, the generated pseudo labels exhibit lower quality than MTrans. Empirical results also demonstrate that IoU-based loss is better than evidential loss to learn the 3D box parameters as the latter treats each 3D box parameter independently.
    
    
\end{itemize}


\subsection{Evidence-aware Multi-task Loss}
\label{edl_regul}

Intuitively, the loss information obtained from several multi-task loss functions during training corresponding to the same object can be utilized to help the model understand the evidence in support to the prediction. In line with the works on uncertainty estimation \cite{Kendall2017WhatUD, Wang2021UncertaintyEF}, we introduce regularized multi-task loss functions based on the form of NLL minimization of aleatoric uncertainty estimation and intuitively aligned with the learned loss attenuation. The main insight behind these loss functions is that the model's predicted evidence is inversely related to the losses for multiple tasks, effectively serving as an evidence regularizer. Let $\mathcal{L}^{'}_{t}$ be the loss function corresponding to the task \textit{t} where $t \in \{$seg, depth, conf, dir$\}$. The proposed evidence-aware multi-task loss is 

\begin{equation}
\label{equation_multiloss}
\begin{aligned}
    \mathcal{L}_{t} &= \dfrac{\mathcal{L}^{'}_t}{1/(\phi -1)} +  \text{ log } \: \dfrac{1}{\phi -1}  \\
    &= (\nu + 2 \alpha -1) \: \mathcal{L}^{'}_t - \text{ log } (\nu + 2 \alpha -1 ),
\end{aligned}
\end{equation}

where $\alpha = |\mathbb{J}|^{-1} \sum_{j \in \mathbb{J}} \alpha_j$ and  $\nu = |\mathbb{J}|^{-1} \sum_{j \in \mathbb{J}} \nu_j$. In Figures \ref{fig:high_epis} and \ref{fig:high_epis_sol}, we demonstrate the effect of including the evidence-aware multi-task loss in the magnitude of the epistemic uncertainties.


\subsection{Uncertainty-aware IoU Loss}

While the NLL-based evidential loss can enable the model to learn the 3D box and evidential parameters, it is not sufficient to regress 3D box variables with a quality comparable to those produced by existing autolabelers. Hence, we propose to include an IoU-based loss inspired by the DIoU loss \cite{Zheng2019DistanceIoULF} similar in form to \ref{equation_multiloss}. This makes $\mathcal{R}$ (penalty term for the prediction and ground truth) and IoU-related term uncertainty-aware. 

\vspace{-0.4cm} 
\begin{equation}
\small
\begin{aligned}
    \mathcal{L}_{IoU} = (\nu + 2 \alpha -1) \cdot (\mathcal{R} + (1-\: IoU))-  \text{ log } (\nu + 2 \alpha -1 ). 
\end{aligned}
\end{equation}

Incorporating this new evidence-aware IoU loss improves the model's ability to handle uncertainty while generating high quality 3D box that is comparable to other 3D autolabelers. In Figure \ref{fig:w_vs_wo_iou_loss}, we demonstrate the effect of including the uncertainty-aware IoU loss in the model performance on the validation set. Note that the original evidence regularizer \cite{Amini2019DeepER} could also train $\gamma_j$. Our empirical findings suggest that eliminating this regularizer is necessary as components of the proposed multi-task losses and the uncertainty-aware IoU loss already serve for regularization purposes. 

\subsection{Training of the 3D Autolabeler } 
\label{edl_fin_training}

\begin{figure*}[!ht]
\captionsetup[subfloat]{labelformat=empty}
    \subfloat[\label{}]{%
       \includegraphics[width=13cm]{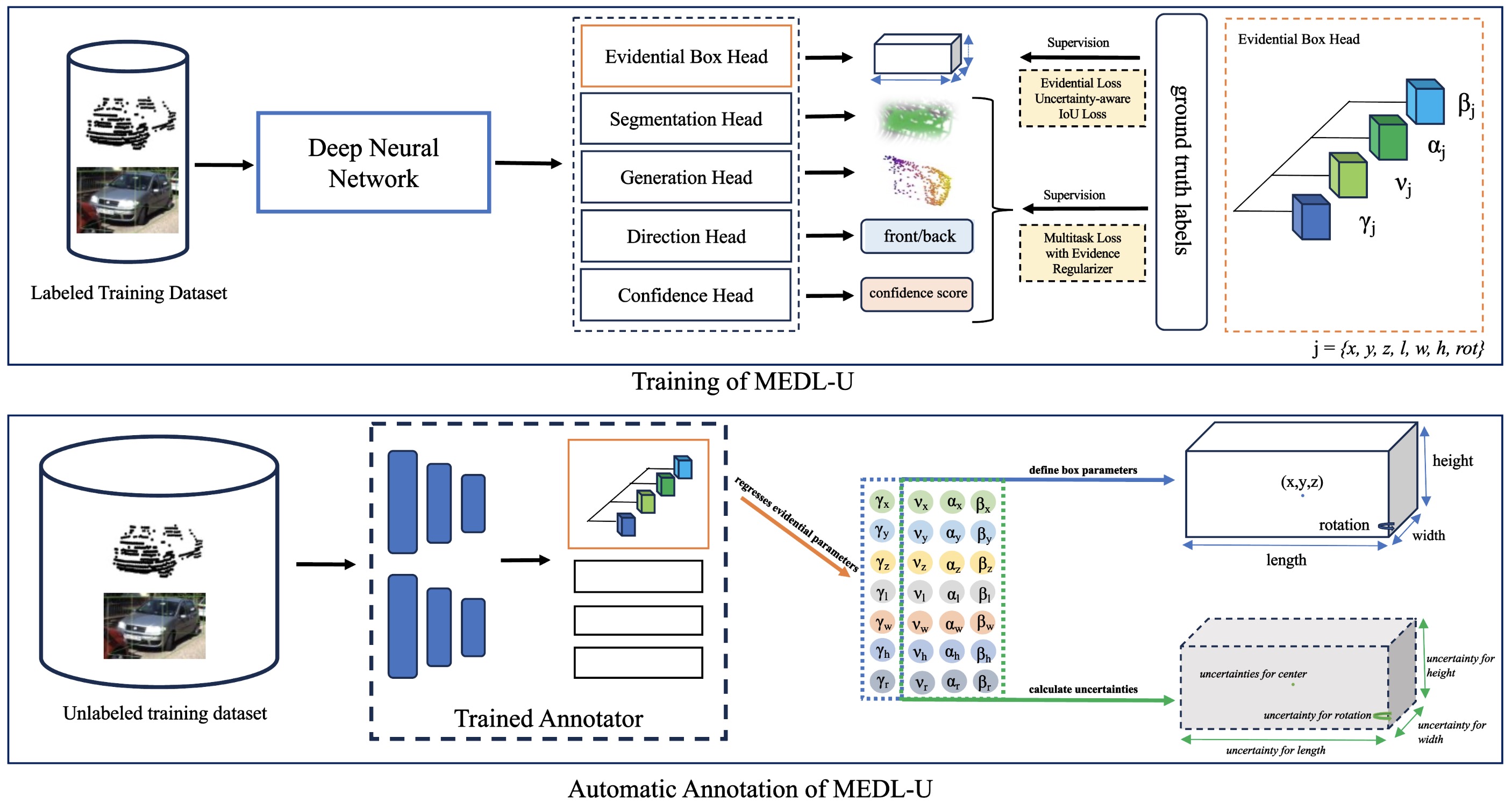}  } 
    \hfill
    \subfloat[\label{}]{%
       \includegraphics[width=5.1cm]{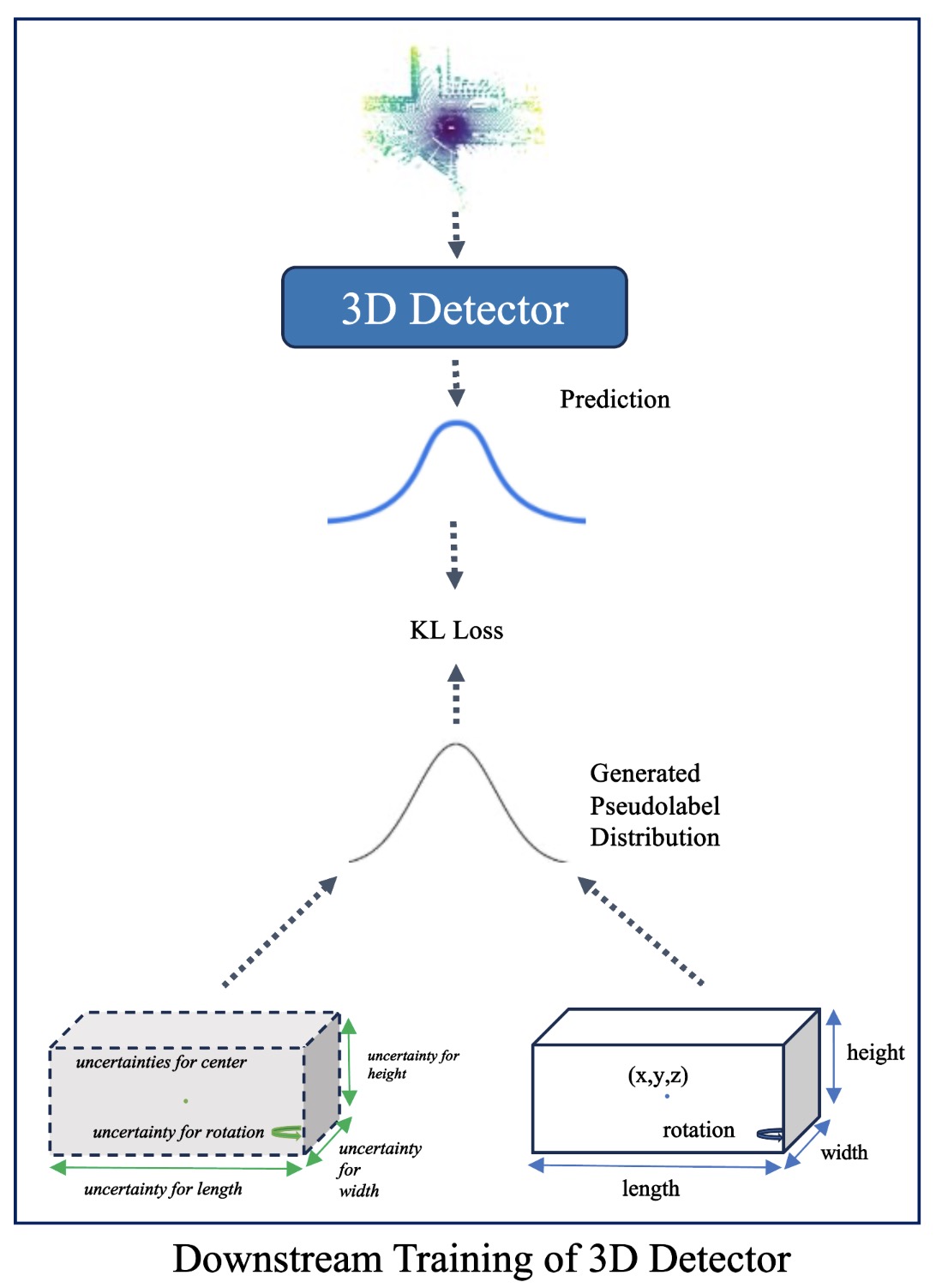}} 
    \vspace*{-\baselineskip} 
    \caption{Architecture of the Training and Automatic Annotation Workflow of MEDL-U. The evidential box head regresses the evidential parameters which can be used to calculate the 3D box parameters and the uncertainties. During the automatic annotation, MEDL-U regresses 3D box parameters and the associated uncertainties for the unlabeled data. In the downstream training of probabilistic 3D detectors, the generated pseudo labels provide supervision during training and the associated box parameter uncertainties serve as factors for reweighting via the KLD loss. \vspace{-0.2cm} }
    \label{fig:method_network}
\vspace*{-\baselineskip} 
\end{figure*}

In summary, the final overall loss function $\mathcal{L}$ is computed as a weighted combination of the evidential loss, the uncertainty-aware IoU loss, and the multi-task losses with evidence regularizers:

\vspace*{-\baselineskip} 
\begin{equation}
\begin{aligned}
    \mathcal{L} = \eta_{seg}\mathcal{L}_{seg} + & \eta_{depth}\mathcal{L}_{depth} +  \eta_{conf}\mathcal{L}_{conf} +  \eta_{dir}\mathcal{L}_{dir} \\ 
    & + \eta_{evi}\mathcal{L}_{evi} + \eta_{IoU}\mathcal{L}_{IoU},
\end{aligned}
\label{eqn:autolabeler_training}
\end{equation}
\vspace*{-\baselineskip} 

where $\eta_{seg}, \eta_{depth}, \eta_{conf}, \eta_{dir}, \eta_{box}$ and $\eta_{evi}$ are hyperparameters. Please refer to Figure \ref{fig:method_network} for the overall workflow of the model.




\subsection{Pseudo Label Uncertainty Post-processing} 
\label{edl_convert}

Prior to utilizing the pseudo labels and uncertainties as supervision signals in the training process of existing 3D detectors, it is necessary to apply a post-processing step to address the variability in the magnitudes of uncertainties associated with each 3D box parameter and to make the uncertainties more appropriate for downstream task. We propose a post-processing procedure that rescales the uncertainties but ensures that the resultant uncertainty values maintain its Spearman's rank correlation with two crucial metrics: the L2 norm of the residuals and the 3D IoU between the predicted box and the ground truth box. Initially, the predicted epistemic uncertainty pertaining to each 3D box parameter $j$, where $j \in \mathbb{J} = \{x, y, z, l, w, h, rot\}$, undergoes a transformation which constrains these uncertainties within the range of 0 to 1 through the application of a simple monotonic function such as min-max scaling. Moreover, we pass uncertainty estimate for each box parameter $x_j$ to a function $f$, formulated as $f(x_j) = x_j^{1/\kappa_j}$, where we carefully select values of $\kappa_j \in [0,10]$ to minimize the NLL of the uncertainties for each 3D box parameter $j$ with respect to the same limited training data used to train the 3D autolabeler. The assumption is that lower NLL means better uncertainty estimates, consequently improving supervision for the downstream task. Lastly, we generate multiple variations of uncertainty by passing them to the function $g(x_j) = x_j^{1/\epsilon_j}$, with $\epsilon_j$ acting as a downstream training hyperparameter. The underlying rationale for introducing $\epsilon_j$ is rooted in the observation that the $\kappa_j$ parameter only effectively minimizes NLL over the limited training data and there must exist an additional parameter which may prove essential for appropriately adjusting the uncertainties to achieve lower NLL over the entire training dataset. 
\vspace{-0.1cm} 

\subsection{Downstream Training via Probabilistic 3D Detector}
\label{prob_3d}

A probabilistic object detector enables the inclusion of pseudo label uncertainties during the training phase, where these uncertainties can be interpreted as factors for reweighting losses. Our work follows \cite{Feng2020LabelsAN, Zhang2022GLENetB3} in transforming a 3D detector from deterministic to probabilistic, where the detection head is enforced to estimate a Gaussian probability distribution over bounding boxes. Let $\theta$ indicate the learnable network weights of a typical detector, $\hat{y}$ indicate the predicted box parameters, and $\hat{\sigma}^2$ the predicted localization variance. Moreover, the pseudo ground truth bounding boxes are also assumed to be Gaussian distributed having a variance $\sigma ^ 2$ where $\sigma ^ 2$ are estimated by MEDL-U. Let the pseudo ground truth bounding box be denoted by $y_g$ and D refer to the pseudo label distribution. Hence, the generated pseudo label uncertainty can be incorporated in the KL Divergence loss between the distribution of prediction and pseudo ground truth in the detection head:
\vspace{-0.3cm} 





\begin{equation}
\begin{aligned}
    L_{reg} &=  D_{KL} (P_{D} (y) || P_{\theta} (y)) \\
    &= log \dfrac{\hat{\sigma}}{\sigma} + \dfrac{\sigma^2}{2 \hat{\sigma} ^2} + \dfrac{(y_{g} - \hat{y})^2}{2 \hat{\sigma} ^ 2}.
\end{aligned}
\end{equation}

Similar to \cite{Zhang2022GLENetB3, He2018BoundingBR}, we also employ 3D Variance Voting which uses the predicted variance $\hat{\sigma}^2$ to combine nearby bounding boxes for better 3D localization prediction.






%% file: p4.exp.tex
\section{Experimental Setup}



\subsection{Dataset}

The KITTI Object Detection dataset \cite{Geiger2012AreWR}, renowned for 3D detection in autonomous driving, is employed in this study. The dataset has a total of 7481 frames with labels in 3D. Following the official procedure, the dataset is divided into training and validation sets, consisting of 3,712 and 3,769 frames, respectively. Similar to previous works \cite{Wei2021FGRFG, Liu2022MAPGenAA, Liu2022MultimodalTF, Qian2023ContextAwareTF}, we concentrate on the Car class and exclude objects with fewer than 5 foreground LiDAR points. 

\subsection{Implementation Details and Model Structure}

Our method is implemented in PyTorch \cite{Paszke2019PyTorchAI}. Similar to the original MTrans \cite{Liu2022MultimodalTF}, MEDL-U architecture incorporates four multimodal self-attention layers, each having a hidden size of 768 and 12 attention heads. Unless otherwise stated, training the autolabeler requires 500 annotated frames only. We employed a dropout rate of 0.4 and utilized the Adam optimizer with a learning rate of $0.60e-04$. Autolabeler training is conducted for 300 epochs, with a batch size of 5. Training of probabilistic 3D detectors is conducted for 80 epochs. Note that only epistemic uncertainties from MEDL-U are utilized. Unless specified differently, hyperparameter tuning on the KITTI validation set suggests using $\epsilon=1$ for PointPillars and CIA-SSD, and $\epsilon=5$ for other detectors. All trainings are executed on NVIDIA RTX 2080Ti GPU.

MEDL-U evidential regression head consists of four output units for each of the seven box attributes. The input to this head are the transformed element representations extracted from the self-attention layer. The evidential regression head comprises a sequence of linear layers, followed by LayerNorm, a Dropout layer, and a ReLU activation function. To ensure that certain values are positive, a Softplus activation is applied to $\nu$, $\alpha$, and $\beta$, where $\alpha$ is then incremented by 1 to ensure $\alpha > 1$. For $\gamma$, a linear activation is used. Overall, MEDL-U has over 23 million trainable parameters. It is trained to learn seven evidential distributions simultaneously for each of the 3D box parameters, along with the other tasks of segmentation, point generation, direction, and confidence prediction. In the next sections, MEDL refers to utilizing the pseudo labels only while MEDL-U refers to utilizing both pseudo labels and the uncertainties in the downstream training of 3D detectors.

\subsection{Evaluation metrics}


\subsubsection{3D Box Prediction}
To assess localization, we measure the Average Precision, specifically for 3D objects ($AP_{3D}$) and Birds Eye View (BEV) with a stringent IoU threshold of 0.70 to determine positive detections. Average Precision at 40 points (R40) means that the precision and recall are calculated at 40 different recall levels.



\subsubsection{Uncertainty Estimation}
Widely employed in previous studies \cite{Gal2015DropoutAA}, the Negative Log-Likelihood (NLL) is utilized as a metric to evaluate the model's ability to estimate uncertainty. Lower NLL values indicate more accurate and more effective uncertainty estimation. Moreover, we also calculate the Spearman's rank correlation coefficients of the predicted uncertainties to the corresponding L2 norm of the residuals.




\subsection{Experiment on Different Kinds of 3D Detectors} 

We evaluate several one-stage and two-stage 3D detectors on the KITTI \textit{val} set when trained using outputs from various annotators. As seen in Table \ref{tab:ap_3d_prob_det_aug}, detectors trained on MEDL-U outputs outperform vanilla deterministic 3D detectors trained on MTrans and MEDL pseudo labels, demonstrating the effectiveness of utilizing not only the pseudo labels but also the uncertainty estimates for each 3D box parameter.


\begin{table}[!b]
\vspace{-0.4cm}
\setlength{\tabcolsep}{2.5pt}
      \scriptsize 
      \centering
      \caption{Comparison of $AP_{3D} \:\: R40$ on the KITTI \textit{val} set using 3D detectors trained on MTrans pseudo labels, MEDL pseudo labels only, and MEDL-U pseudo labels and uncertainties. Results are produced from our own experiments. }
      \begin{tabular}{|l|c c c|c c c| c c c|}
      \hline
         & \multicolumn{3}{c|} 
         {\scriptsize{Easy}} &\multicolumn{3}{c|}{\scriptsize{Moderate}} &\multicolumn{3}{c|}{\scriptsize{Hard}} \\
         \hline
         \scriptsize{Detector} & {\tiny MTrans} & {\tiny MEDL} &  {\tiny MEDL-U} & {\tiny MTrans} & {\tiny MEDL} &  {\tiny MEDL-U} & {\tiny MTrans} & {\tiny MEDL} &  {\tiny MEDL-U} \\    
         \hline 
         {\scriptsize SECOND \cite{Yan2018SECONDSE}}     &90.69&89.53&\textbf{90.95}     &79.63&80.53&\textbf{82.69}      &76.11&76.20&\textbf{77.82}   \\   
        {\scriptsize PointPillars \cite{Lang2018PointPillarsFE}}  &86.70&87.34&\textbf{91.16}     &75.36&78.07&\textbf{80.44}      &72.06&74.45&\textbf{75.43}    \\  
        {\scriptsize CIA-SSD \cite{Zheng2020CIASSDCI}}    &89.81&90.50&\textbf{92.33}     &78.52&80.79&\textbf{83.27}      &73.22&75.48&\textbf{78.30}    \\  
        {\scriptsize Voxel RCNN \cite{Deng2020VoxelRT}}   &92.25&92.30&\textbf{92.59}     &83.25&82.95&\textbf{83.72}      &80.11&79.92&\textbf{80.73}    \\   
        {\scriptsize PointRCNN \cite{Shi2018PointRCNN3O}}    &91.67&91.56&\textbf{92.47}     &80.73&80.45&\textbf{81.59}      &75.75&77.54&\textbf{78.68}    \\   
        \hline
      \end{tabular}
      
      \label{tab:ap_3d_prob_det_aug}

\end{table}

\vspace{-0.05cm}

\subsection{Comparison with 3D Automatic Annotation Frameworks} 

As shown in Table \ref{val_methods_comparison}, evaluating the probabilistic PointRCNN trained with MEDL-U ouputs on the KITTI \textit{val} set yields superior performance relative to all existing and current 3D autolabeling methods in terms of $AP_{3D}$. 

\begin{table}[t!]
\setlength{\tabcolsep}{4pt}
\scriptsize
    \centering
    \caption{$AP_{3D}$ Results on KITTI \textit{val} set, compared to the fully supervised PointRCNN and other weakly supervised baselines. Results here are from the official published results.}
    \begin{tabular}{| l | c | c | c c c|} 
    \hline
    Method & Reference & Full Supervision & Easy & Moderate & Hard  \\ 
    \hline
    PointRCNN \cite{Shi2018PointRCNN3O} & CVPR 2019 & \checkmark & 88.99 & 78.71 & 78.21 \\ 
    \hline
    WS3D \cite{Meng2020WeaklyS3} & ECCV 2020 & BEV Centroid &  84.04 & 75.10 & 73.29 \\ 
    WS3D (2021) \cite{Meng2021TowardsAW} & TPAMI 2022 & BEV Centroid &  85.04 & 75.94 & 74.38 \\ 
    FGR \cite{Wei2021FGRFG} & ICRA 2021 & 2D Box &  86.68 & 73.55 & 67.91 \\ 
    MAP-Gen \cite{Liu2022MAPGenAA} & ICPR 2022 & 2D Box &  87.87 & 77.98 & 76.18 \\ 
    MTrans \cite{Liu2022MultimodalTF} & ECCV 2022 & 2D Box &  88.72 & 78.84 & 77.43 \\ 
    CAT \cite{Qian2023ContextAwareTF} & AAAI 2023 & 2D Box &  89.19 & 79.02 & 77.74\\ 
    MEDL (Ours) & - & 2D Box &   89.07 & 78.68 & 76.99\\ 
    MEDL-U (Ours) & -  & 2D Box & \textbf{89.26} & \textbf{79.27} & \textbf{78.05} \\ 
    \hline 
    \end{tabular}
    
    \label{val_methods_comparison}
\end{table}

In Table \ref{test_methods_comparison}, probabilistic PointRCNN trained with outputs of MEDL-U on the entire KITTI training and val sets results in better performance on the KITTI official \textit{test} set compared to PointRCNN trained with vanilla MTrans. Moreover, PointRCNN trained with MEDL-U outputs yields superior performance in terms of $AP_{3D}$ and $AP_{BEV}$ for both Easy and Moderate levels relative to all existing 3D automatic labeling method. MEDL-U does not outperform CAT across all difficulty levels, which is understandable considering that MEDL-U is built upon MTrans, chosen for its open-source availability. Moreover, CAT is trained for 1000 epochs and a batch size of 24, which is different from the training setting for MTrans and MEDL-U. We argue that the enhancements seen in MEDL-U over MTrans can also be applied to CAT.


\begin{table}[t!]
\vspace{-0.1cm}
\setlength{\tabcolsep}{2pt}
\scriptsize
    \centering
    \caption{Results of KITTI official \textit{test} set, compared to the fully supervised PointRCNN and other weakly supervised baselines. Results here are from official published results.}
    \begin{tabular}{| l | c |  c c c | c c c|}
    \hline
    & & \multicolumn{3}{c}{$AP_{3D}$} & \multicolumn{3}{|c|}{$AP_{BEV}$}  \\
    \hline
    Method & Modality  & Easy & Moderate & Hard & \:\:\:Easy\:\:\: & Moderate & Hard  \\ 
    \hline
    PointRCNN \cite{Shi2018PointRCNN3O} &LiDAR & 86.96 & 75.64 & 70.70 & 92.13 & 87.39 & 82.72 \\ 
    \hline
    WS3D \cite{Meng2020WeaklyS3}  & LiDAR  & 80.15 & 75.22 & 70.05 & 90.11 & 84.02 & 76.97 \\ 
    WS3D (2021) \cite{Meng2021TowardsAW} & LiDAR  & 80.99 & 70.59 & 64.23 & 90.96 & 84.93 & 77.96  \\ 
    FGR \cite{Wei2021FGRFG} & LiDAR  & 80.26 & 68.47 & 61.57 & 90.64 & 82.67 & 75.46 \\ 
    MAP-Gen \cite{Liu2022MAPGenAA} & LiDAR+RGB  & 81.51 & 74.14 & 67.55 & 90.61 & 85.91 & 80.58  \\ 
    MTrans \cite{Liu2022MultimodalTF} & LiDAR+RGB  & 83.42 & 75.07 & 68.26 & 91.42 & 85.96 & 78.82 \\ 
    CAT \cite{Qian2023ContextAwareTF} & LiDAR  & 84.84 & 75.22 & \textbf{70.05} & 91.48 & 85.97 & \textbf{80.93} \\ 
    MEDL-U (Ours) & LiDAR+RGB  & \textbf{85.49}  &	\textbf{75.96} &	69.12  & \textbf{91.86} &	\textbf{86.68} & 79.44\\ 
    \hline 
    \end{tabular}
    \label{test_methods_comparison}
\end{table}



We also show evaluation performance on the KITTI \textit{val} and \textit{test} set using PointPillars when MTrans and MEDL-U are trained with 500 and 125 annotated frames. MEDL-U significantly improves the baseline as shown in Table \ref{val_medlu_vs_mtrans_500_125}.


\begin{table}[t!]
\setlength{\tabcolsep}{2pt}
\vspace{-0.2cm}
\scriptsize
    \centering
    \caption{$AP_{3D}$ Results on KITTI \textit{val} and official \textit{test} set using PointPillars when the 3D autolabelers are trained using 500 and 125 frames of annotated data. We produce KITTI \textit{test} results for MTrans + PointPillars (125f), while other results on MTrans are officially published results.}
    \begin{tabular}{| l | c | c c c | c c c|}
    \hline
    & &\multicolumn{3}{c|}{KITTI \textit{val} set} &\multicolumn{3}{c|}{KITTI official \textit{test} set} \\
    \hline 
    Method & Supervision & Easy & Moderate & Hard & Easy & Moderate & Hard \\ 
    \hline
    MTrans + PointPillars &  500f & 86.69 & 76.56 & 72.38 & 77.65 & 67.48 & 62.38 \\ 
    MEDL-U + PointPillars &  500f &  \textbf{88.12} & \textbf{78.24} & \textbf{76.58} & \textbf{82.77}  & \textbf{72.74} &	\textbf{65.57}  \\ 
    \hline
    MTrans + PointPillars &  125f & 83.70 & 71.66 & 66.67  & 71.49  & 59.04 & 51.79  \\ 
    MEDL-U + PointPillars &  125f & \textbf{85.49} & \textbf{74.66} & \textbf{66.95} & \textbf{78.13} & \textbf{66.46} & \textbf{57.40}  \\ 
    \hline 
    \end{tabular}
    \vspace{-0.3cm}
    \label{val_medlu_vs_mtrans_500_125}
\end{table}



\vspace{-0.1cm}
\subsection{Comparison with Other Uncertainty Estimation Methods}

 Using 3D box parameter uncertainties generated by MEDL-U and other popular uncertainty estimation methods, we evaluate probabilistic version of PointRCNN on the KITTI \textit{val} set. Three baseline methods were implemented: (1) A Monte Carlo dropout (MC Dropout) system with a dropout rate of 0.2 and was forwarded 5 times during inference. (2) A Deep Ensemble of 5 systems trained with different random seeds. (3) Confidence score predicted by vanilla MTrans was used as proxy to uncertainty. As shown in Table \ref{tab:ours_vs_unc_methods_ap3dr40}, PointRCNN trained with MEDL-U ($\epsilon=5$) yields the overall best result in terms of $AP_{3D} \: R40$. Noticeably, using other uncertainty estimation methods to generate uncertainties also effectively increase $AP_{3D} \: R40$ of the \textit{base} method, although MC Dropout and Deep Ensemble come at the cost of huge additional computational overhead. While the Deep Ensemble approach can improve the \textit{base} result, the original pseudo label quality is relatively poor.

\begin{table}[b!]
\scriptsize
    \centering
    \setlength{\tabcolsep}{8pt}
    \vspace{-0.3cm}
    \caption{Comparison of $AP_{3D}$ $R40$ on KITTI \textit{val} set when PointRCNN is trained using outputs from different uncertainty estimation methods. \textit{base} refers to utilizing pseudo labels only in training PointRCNN.} 
    \begin{tabular}{|l|c|c|c|} %
        \hline
        Method &  Easy & Moderate & Hard \\ 
        \hline
        vanilla MTrans \cite{Liu2022MultimodalTF} & 91.67    & 80.73 & 75.75  \\  
        conf ($\epsilon = 4$) & 92.10 & 80.87 & 75.91 \\ 
        conf ($\epsilon = 5$) & 92.59 & 81.42 & 76.41 \\ 
        \hline
        MC Dropout (\textit{base}) & 92.14  & 80.43 & 75.76                   \\ 
        MC Dropout ($\epsilon=4$) &  92.50  & 81.22 & 76.44             \\ 
        MC Dropout ($\epsilon$ = 5) &  \textbf{92.91}  & 81.35 & 76.50           \\ 
        \hline
        Deep Ensemble (\textit{base}) & 84.47   & 74.99 & 72.14  \\ 
        Deep Ensemble ($\epsilon=4$) & 85.06  & 75.99 & 71.18         \\ %
        Deep Ensemble ($\epsilon=5$) & 84.78  & 75.49 & 72.82        \\ %
        \hline 
        MEDL (\textit{base}) & 91.56 & 80.45 & 77.54 \\ %
        MEDL-U ($\epsilon = 4$) & 91.73 & 81.11 & 78.27 \\ %
        MEDL-U ($\epsilon$ = 5) & 92.47 & \textbf{81.59} & \textbf{78.68}   \\  %
        \hline
    \end{tabular}
    
    \label{tab:ours_vs_unc_methods_ap3dr40}
\end{table}




\subsection{Uncertainty Evaluation and Ablation Results}

Table \ref{tab:unc_eval} shows that the uncertainties generated by MEDL-U achieve the overall best result in terms of NLL and Spearman's rank correlation coefficient. Specifically, it achieves the highest uncertainty Spearman's rank correlation with residuals for the $x$ and $y$ coordinate of the center, height, and rotation, and consistently among the two lowest calculated NLL for all the 3D box parameters. Details on ablation results and results on varying $\epsilon$ for the downstream task can be seen in Tables \ref{ablation_corr} and \ref{tab:eps_ablation}, respectively. 

 \begin{table}[t!]
 \setlength{\tabcolsep}{2.5pt}
     \scriptsize
      \centering
      \caption{Evaluating uncertainties using NLL and Spearman's rank correlation with residuals. Values in bold achieve the best score. Underlined values achieve the second best.} 
      \begin{tabular}{|l|c c c c c c c | c c c c c c c|}
          \hline
          & \multicolumn{7}{c|}{Negative Log Likelihood (NLL)} & \multicolumn{7}{c|}{Correlation Coefficient (in \%)} \\
          \hline 
          & x&y&z&l&w&h&rot & x&y&z&l&w&h&rot\\ 
         \hline 
          Confidence score &4.5&2.4&\textbf{-1.9}&3.6&1.4&\textbf{-1.3}&5.3 & \underline{40}&50&33&\textbf{44}&\underline{43}&35&25  \\  
         
         MC Dropout & \underline{2.2}&1.7&-1.3&1.9&1.4&-0.6&5.5 & 39&45&\underline{45}&-2&\textbf{45}&21&24   \\   

         Deep Ensemble &4.4&\textbf{0.1}&-1.5&\textbf{0.6}&\textbf{-1.2}&-0.7&\textbf{0.9} & \textbf{49}&\underline{55}&\textbf{49}&\underline{41}&37&\underline{39}&\underline{31}  \\
         MEDL-U & \textbf{1.9}&\underline{0.4}&\underline{-1.8}&\underline{1.0}&\underline{-1.0}&\underline{-1.2}&\underline{1.3} & \textbf{49}&\textbf{56}&43&39&39&\textbf{40}&\textbf{61}\\
         \hline 
      \end{tabular}
    
    \label{tab:unc_eval}
\end{table}

\begin{table}[t!]
\setlength{\tabcolsep}{1.5pt}
\vspace{-0.2cm}
\scriptsize
    \centering
    \caption{Ablation Results showing Spearman's rank correlation (in \%) between generated uncertainties and the L2 norm of the residuals.}
    \begin{tabular}{| l | c | c  | c | c | c  | c |c|}
    \hline
    Model Design & \textit{x} & \textit{y} & \textit{z} & \textit{l} & \textit{w} & \textit{h} & \textit{rot} \\ 
    \hline
    Replace original IoU loss with $\mathcal{L}_{evi}$ \& $\mathcal{L}_R$ & 45.8 & 54.8 & 41.2 & 35.6 & 35.2 & 39.9 & 59.3 \\ 
    Include evidence-aware multi-task loss $\mathcal{L}_{t}$ & 47.7 & 55.4 & 42.3 & 36.1 & 36.7 & 39.7 & 59.8 \\
    Replace $\mathcal{L}_R$ with uncertainty-aware IoU loss $\mathcal{L}_{IoU}$ & \textbf{49.0} & \textbf{55.8} & \textbf{42.8} & \textbf{38.7} & \textbf{38.6} & \textbf{39.9} & \textbf{60.8} \\
    \hline 
    \end{tabular}
    \label{ablation_corr}
\end{table}

\begin{table}[t!]
    \setlength{\tabcolsep}{4pt}
    \renewcommand{\arraystretch}{1}
    \vspace{-0.2cm}
    \scriptsize
    \centering
    \caption{Comparison of the $AP_{3D}$ $R40$ when training Voxel R-CNN using different values of $\epsilon$.} 
        \begin{tabular}{|l|c|c|c|c|c|c|} 
            \hline
             & \multicolumn{6}{c|}{Voxel R-CNN + MEDL-U} \\
            \hline
            $AP_{3D} \: R40$ & 1 & 2 & 3 & 4 & 5 & 6 \\ 
            \hline 
            Easy &  92.32 & 92.15 & 92.15  & 92.16 & \textbf{92.59} & 92.15 \\
            Moderate & 83.54 & 83.35 & 83.50 & 83.48 & \textbf{83.72} & 83.51 \\
            Hard & 78.65 & 80.39 & 80.59 & 80.60 & \textbf{80.73} & 80.52 \\
            \hline
        \end{tabular}
        \label{tab:eps_ablation}
        \vspace{-0.5cm}
\end{table}

%% file: p5.con.tex
\section{Conclusion}

In this paper, we propose MEDL-U, a 3D automatic labeler with EDL framework that generates not only high-quality pseudo labels but also uncertainties associated with each 3D box parameter. Compared with previous autolabeling approaches, our method achieves overall better results in the downstream 3D object detection task on the KITTI \textit{val} and \textit{test} set, showing the importance of quantifying uncertainties or noise in pseudo labels.

%% file: appendix.tex
\subsection{Details of MTrans}

The proposed framework MEDL-U is built from MTrans \cite{Liu2022MultimodalTF}. In this section, we briefly describe the architecture of MTrans. MTrans generates object features from the multimodal image and point cloud inputs. Then these features are encoded to generate embedding vectors using sinusoidal position embedding for 2D pixels, MLP for 3D coordinates, and 3-layer CNN for the image patch features; these features are fused and passed to a fully-connected layer to generate element representations. Then a multimodal self-attention mechanism manipulates and transforms these features to generate object-level representations. For the multi-task supervision, multiple prediction heads take the transformed representation as input. In the self-attention mechanism, MTrans enriches the sparse point cloud by defining a training mechanism for context points with known 3D coordinates, padding points which are randomly dropped points, and target points which have a corresponding uniformly distributed image pixels. The various heads are the 2-layer MLP segmentation head, 2-layer MLP point generation head, box regression head, binary direction head (front or back), and the confidence head. Note that MTrans only included the confidence head for the purpose of converting the autolabeler to a 3D object detector version. But in our framework, we included the confidence head so that we may use the confidence estimates as a proxy to uncertainty, which can be compared with our proposed uncertainty estimation approach. 

To supervise the training of these multiple heads, sum of cross-entropy and dice loss is used for the binary segmentation task loss $\mathcal{L}_{seg}$, smooth $L_1$ loss for the point generation task loss $\mathcal{L}_{depth}$ with a self-supervised mask-and-predict strategy, dIoU loss for the box loss, cross-entropy loss $\mathcal{L}_{dir}$ for the binary direction classification, and smooth $L_1$ loss $\mathcal{L}_{conf}$ for the confidence prediction. We advise the reader to read \cite{Liu2022MultimodalTF} for a more complete information about MTrans.

\subsection{Additional Implementation Details}

Similar to the configuration of MTrans, we apply the standard image augmentation methods of auto contrast, random sharpness, and color jittering as well as point cloud augmentation methods of random translation, scaling and mirroring. For each object, the cloud size is set to 512 for context plus padding points, and there are 784 image-sampled target points. For the training of the autolabeler, we set the following loss weights for \ref{eqn:autolabeler_training}: $\eta_{seg} = 1$, $\eta_{depth} = 1$, $\eta_{conf} = 1$, $\eta_{dir} = 1$, $\eta_{evi} = 2$, and $\eta_{IoU}=5$.

Moreover, in MTrans, the ranges of the predicted and ground truth yaw are limited to $[-\pi/2, \pi/2]$ and $[-\pi$ to $\pi]$, respectively. Hence, if we were to employ the evidential loss to supervise the predicted yaw, this would lead to inconsistencies, since a component of the evidential loss involves a direct calculation of the difference between the predicted yaw and the ground truth yaw. As a result, the constraints imposed on the predicted yaw in vanilla MTrans may not be compatible with the requirements of the evidential loss. Different from MTrans, we wrap the ground truth yaw Euler angles to the range of [-$\pi$/2, $\pi$/2] to make it appropriate for the evidential loss corresponding to rotation. Moreover, the presence of the direction loss is sufficient for accurate yaw prediction. By wrapping the yaw Euler angle to the range [-$\pi$/2, $\pi$/2], we ensure that the ground truth yaw angle stays within a compact interval, which makes it appropriate for the evidential loss corresponding to rotation.


\subsection{Effect of the Evidence-aware Multi-task Loss}
\setcounter{figure}{0}

\begin{figure}[!ht]
    \centering
    \begin{subfigure}{0.495\textwidth}
        \centering
        \includegraphics[width=\linewidth]{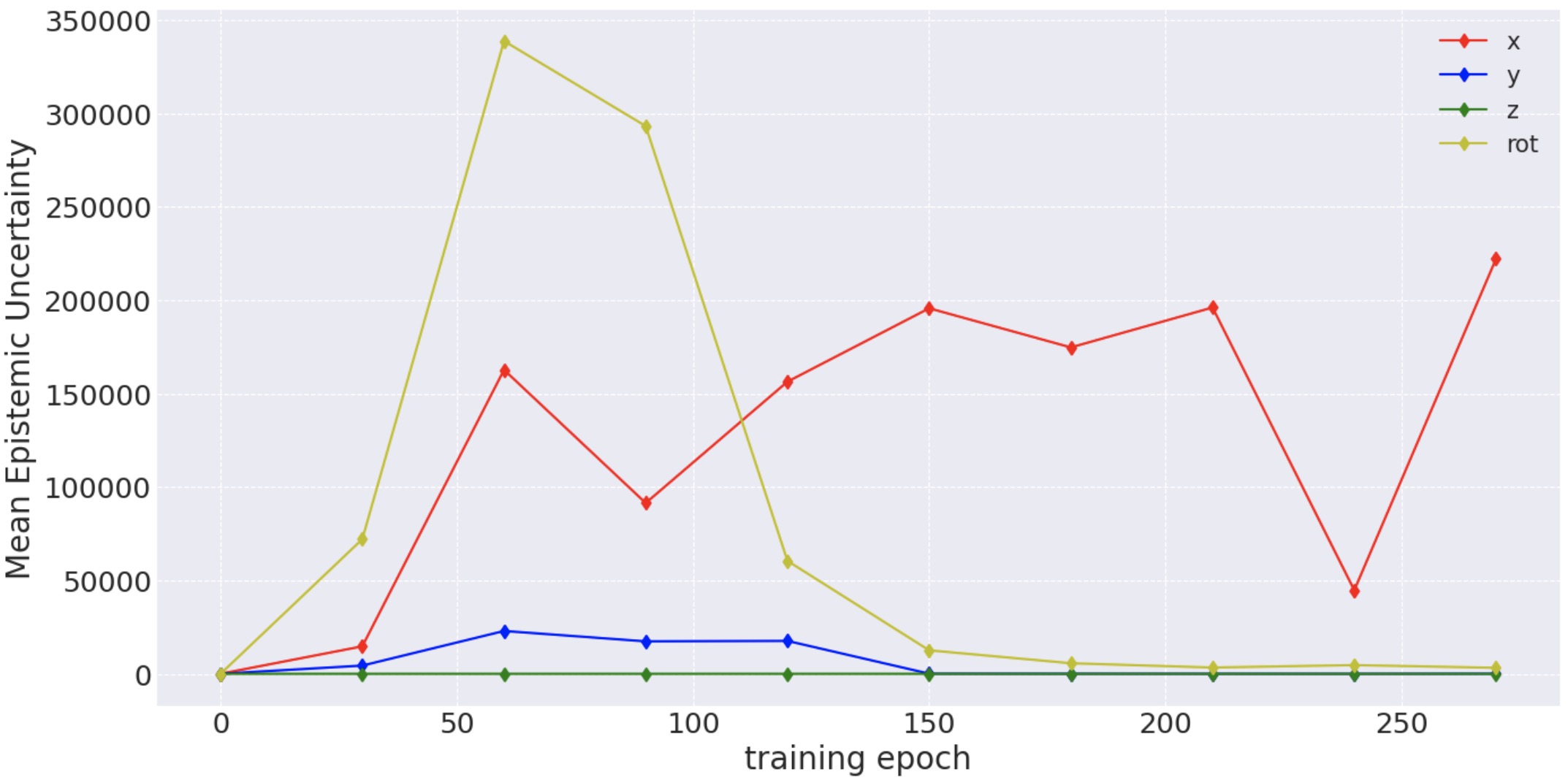}
    \end{subfigure}
    \vspace{0.10cm}
    \begin{subfigure}{0.495\textwidth}
        \centering
        \includegraphics[width=\linewidth]{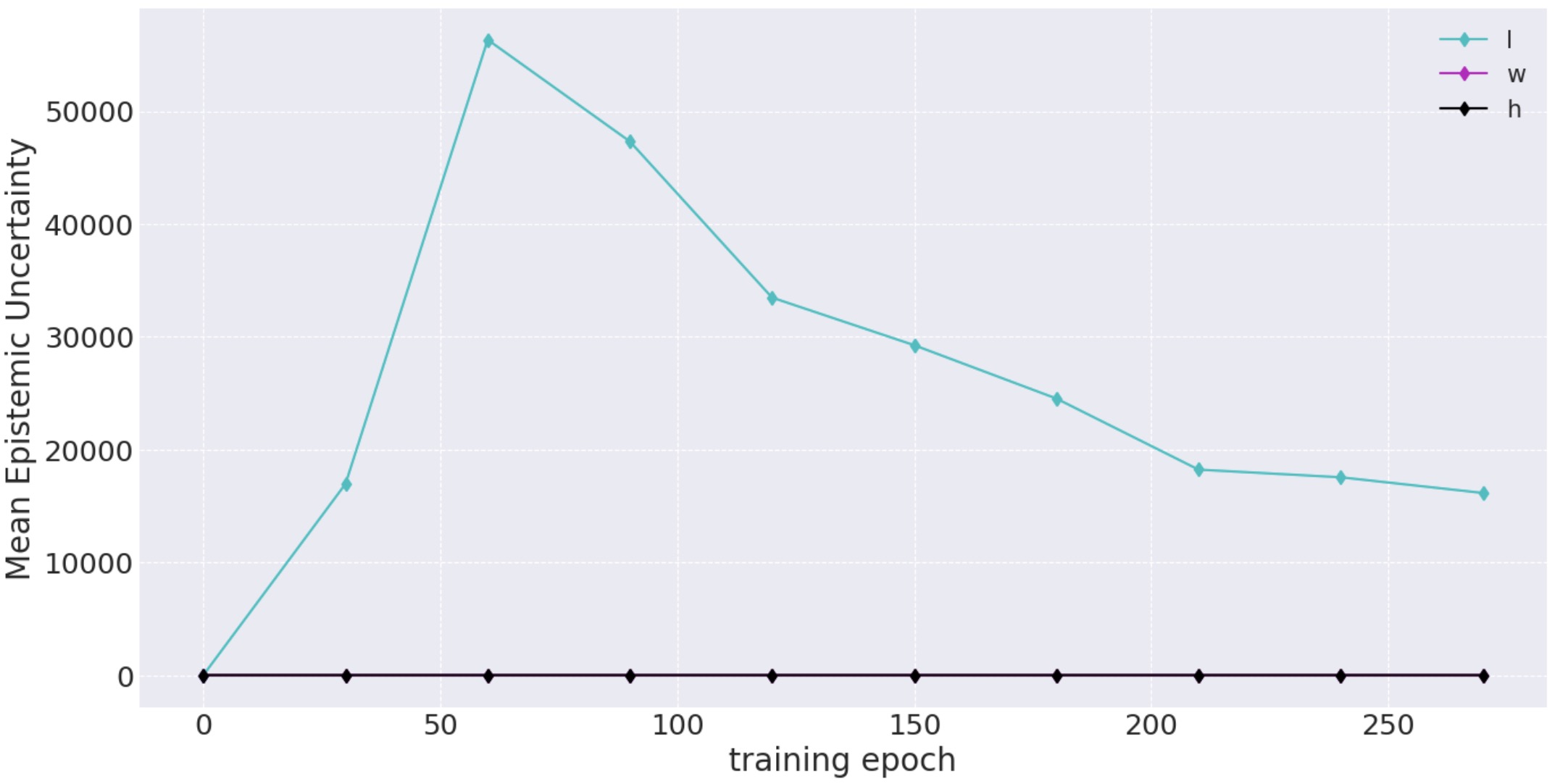}
    \end{subfigure}
    \caption{Visualizing the mean epistemic uncertainties for each 3D box parameter during training when the model only uses the evidential loss with uncertainty calibration term \cite{Amini2019DeepER} to supervise the training of the 3D autolabeler.}
    \label{fig:high_epis}
\end{figure}

\begin{figure}[!ht]
    \centering
    \begin{subfigure}{0.495\textwidth}
        \centering
        \includegraphics[width=\linewidth]{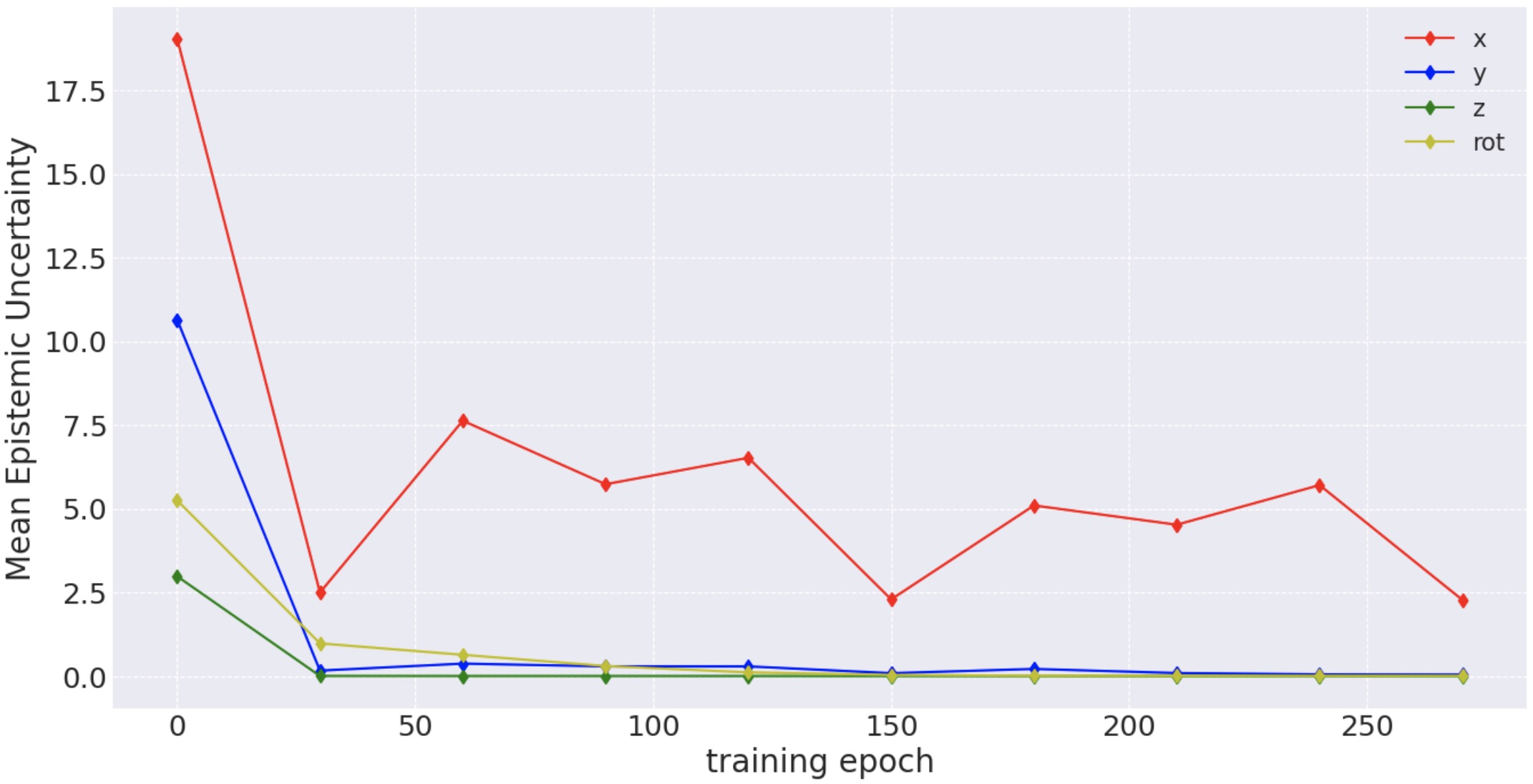}
    \end{subfigure}
    \vspace{0.10cm}
    \begin{subfigure}{0.495\textwidth}
        \centering
        \includegraphics[width=\linewidth]{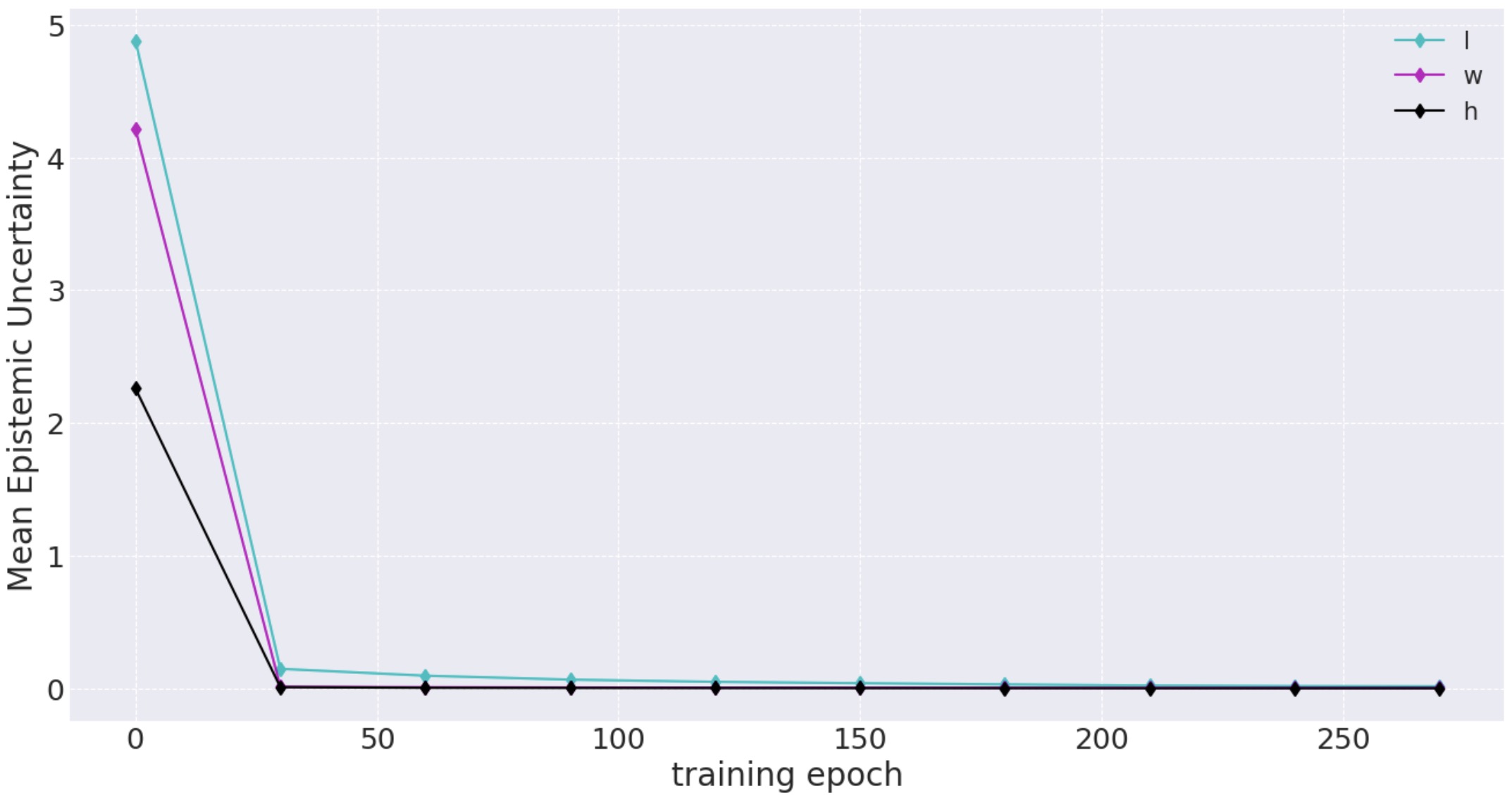}
    \end{subfigure}
    \caption{Visualizing the mean epistemic uncertainties for each 3D box parameter during training when the model includes the evidence-aware multi-task loss to supervise the training of the 3D autolabeler.}
    \label{fig:high_epis_sol}
\end{figure}

Figure \ref{fig:high_epis} visualizes the excessively high epistemic uncertainties generated for the 3D box parameters during training and the huge variations in the magnitudes of the uncertainty for each 3D box parameter when only the evidential loss with uncertainty calibration term is used to supervise the model training. Figures \ref{fig:high_epis_sol} shows that including the evidence-aware multi-task loss during training mitigates the problem of generating excessively high epistemic uncertainties and reduces the differences of the magnitude of the epistemic uncertainties for each 3D box parameter. Moreover, the trend of the mean epistemic uncertainties during training is generally downward, which is indicative of the model being less uncertain as the training progresses.

\subsection{Effect of the Uncertainty-aware IoU Loss}

\begin{figure}[!ht]
    \centering
        \includegraphics[width=\linewidth]{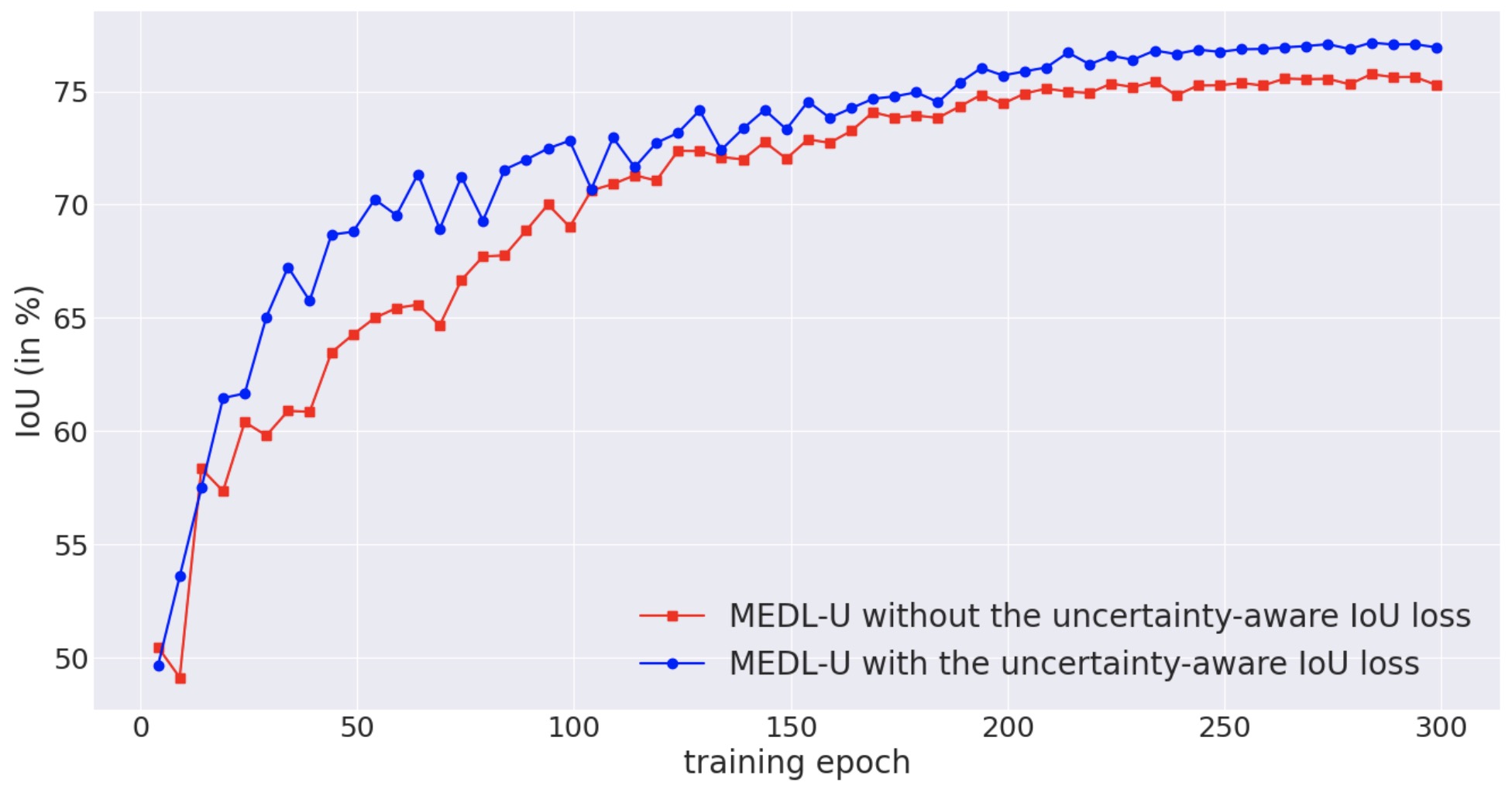}
        \caption{Visualizing the effect on IoU of prediction and the ground truth after incorporating the uncertainty-aware IoU loss on the model during training on the KITTI \textit{val} set.}
    \label{fig:w_vs_wo_iou_loss}
\end{figure}

In Figure \ref{fig:w_vs_wo_iou_loss}, we can see an improved IoU of the predictions with the ground truth during training on the validation set because of the inclusion of the uncertainty-aware IoU loss. 

\subsection{Analysis of the Uncertainties}


In this section, we demonstrate that MEDL-U can generate reasonable uncertainty estimates (both aleatoric and epistemic uncertainties) for each of the 3D box parameter $j \in \mathbb{J} = \{x, \: y, \:z, \:l, \:w, \:h, rot\}$, which could give an intuition as to why using the information on pseudo label uncertainty improves the performance of the downstream 3D detection task.

\subsubsection{Spearman's Rank Correlation of Epistemic Uncertainties with the Prediction 3D IoU with GT Box}

\begin{figure}[!ht]
    \centering
    \begin{subfigure}{0.49\textwidth}
        \centering
        \includegraphics[width=\linewidth]{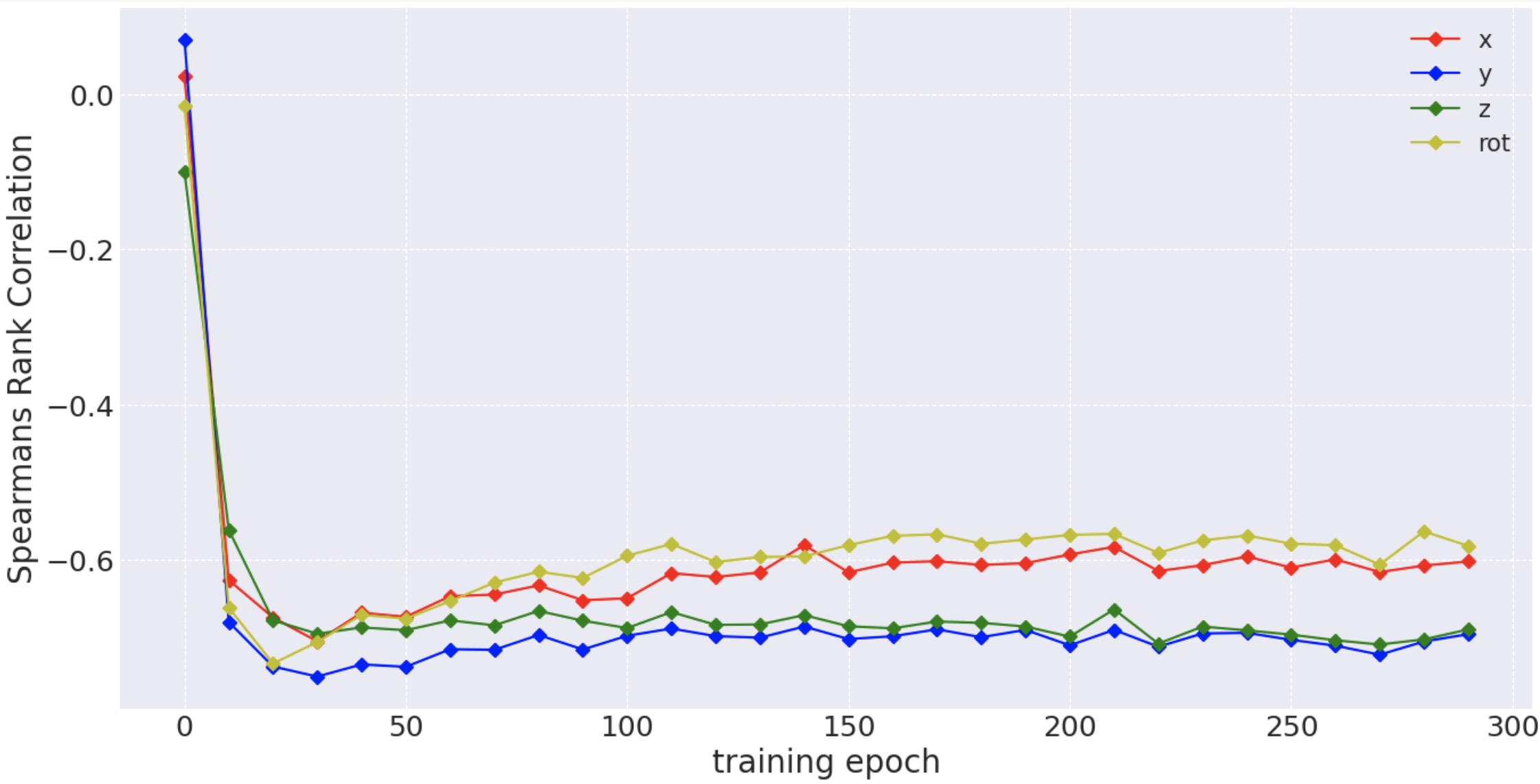}
        \caption{Progression of the Spearman's Rank Correlation Coefficient of the Epistemic Uncertainties for each 3D Box Parameter \textit{x,y,z,} and \textit{rot} and the 3D Box IoU to GT.}
        \label{fig:corr_iou_epis_per_dim}
    \end{subfigure}
    \hfill
    \begin{subfigure}{0.49\textwidth}
        \centering
        \includegraphics[width=\linewidth]{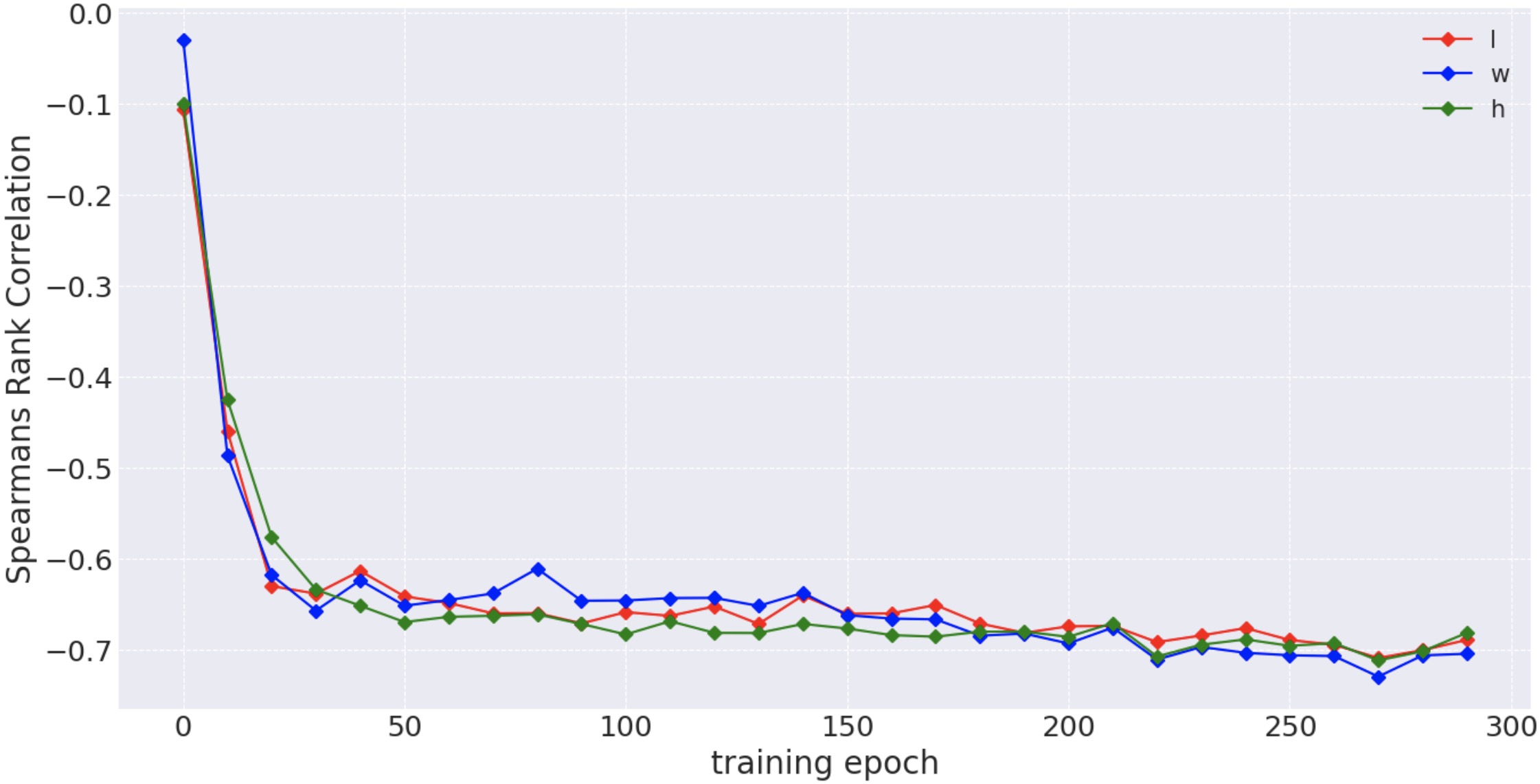}
        \caption{Progression of the Spearman's Rank Correlation Coefficient of the Epistemic Uncertainties for each 3D Box Parameter \textit{l, w, } and \textit{h} and the 3D Box IoU to GT.}
        \label{fig:corr_spr_iou_epis_per_dim}
    \end{subfigure}
    \caption{Visualization of the Spearman's rank correlation Coefficient of the epistemic uncertainties for each 3D box parameter and the GT 3D IoU.}
    \label{fig:corr_iou_epis_per_dim_total}
\end{figure}

Figure \ref{fig:corr_iou_epis_per_dim_total} shows that during training, the progression of the correlation for the length, width, height, and the z-coordinate of the center is decreasing, while the progression of the correlation for the rotation, x-coordinate, and y-coordinate of the center has plateaued after a few training epochs. However, after training, the correlation of epistemic uncertainties of each 3D box parameter is negatively correlated with the GT IoU (negative correlation with magnitude greater than 55
\%). Recall that epistemic uncertainties deal with model uncertainties. This can be intuitively interpreted as the model becoming more confident of its predictions during the training process.

\subsubsection{Spearman's Rank Correlation of Epistemic Uncertainties with the $L2$ Norm of the Residuals}

\begin{figure}[!ht]
    \centering
    \begin{subfigure}{0.49\textwidth}
        \centering
        \includegraphics[width=\linewidth]{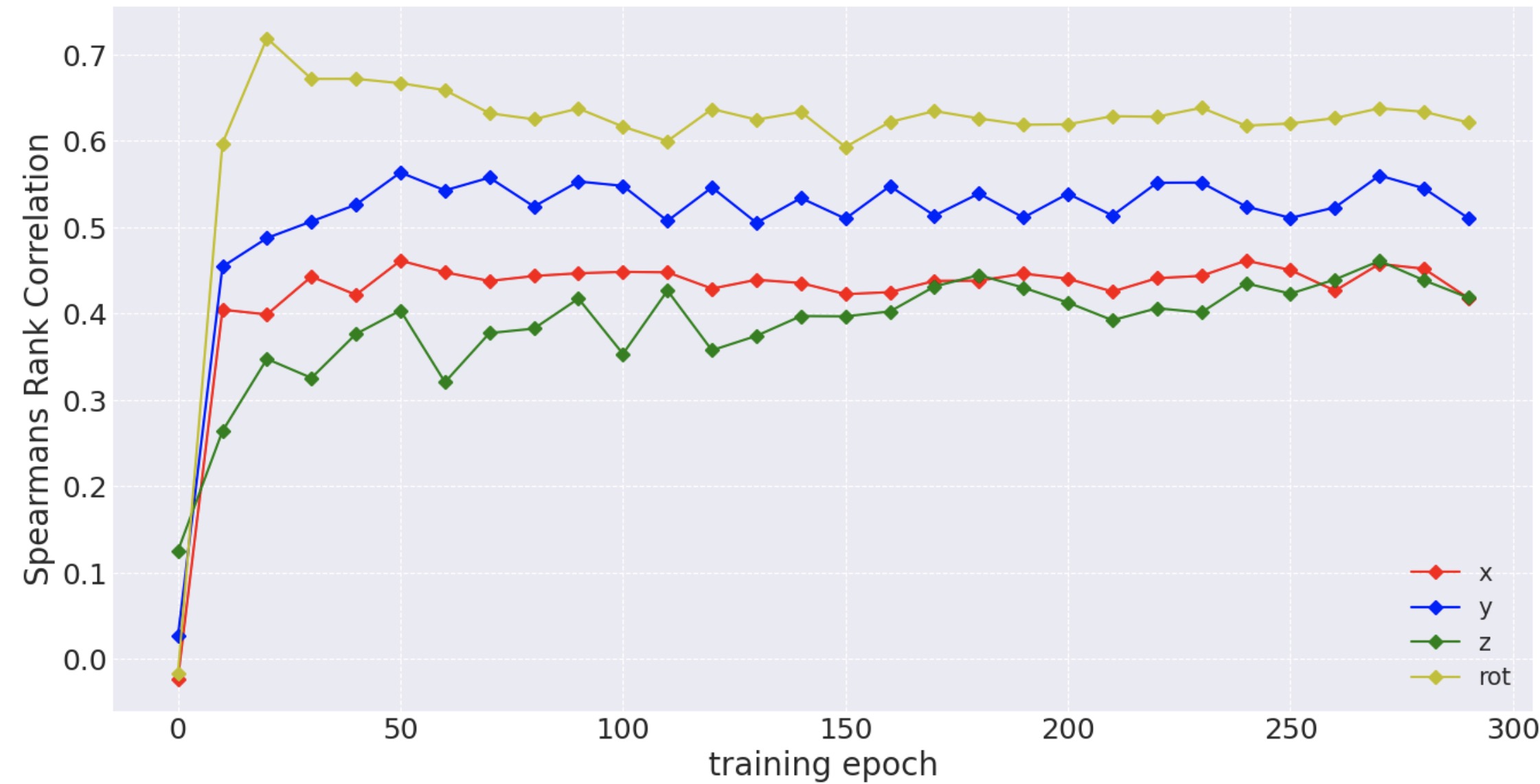}
        \caption{Progression of the Spearman's rank correlation Coefficient of the Epistemic Uncertainties for Each 3D Box Parameter \textit{x,y,z,} and \textit{rot} and the L2 Norm of the Residuals.}
        \label{fig:corr_spr_l2res_alea}
    \end{subfigure}
    \hfill
    \begin{subfigure}{0.49\textwidth}
        \centering
        \includegraphics[width=\linewidth]{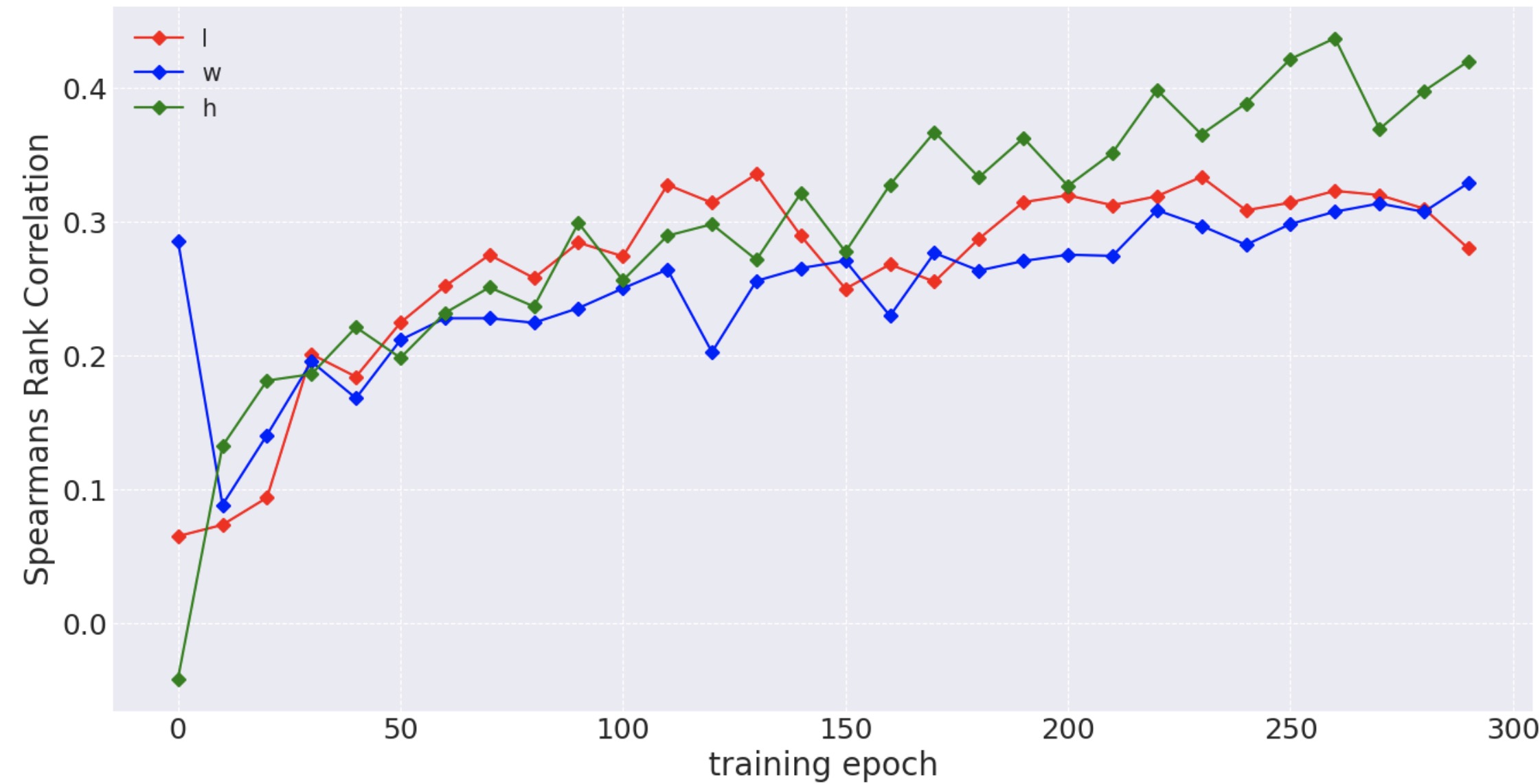}
        \caption{Progression of the Spearman's Rank correlation Coefficient of the Epistemic Uncertainties for Each 3D Box Parameter \textit{l,w,} and \textit{h} and the L2 Norm of the Residuals.}
        \label{fig:corr_spr_l2res_epis}
    \end{subfigure}
    \caption{Visualization of the Spearman's rank correlation Coefficient of the Epistemic Uncertainties and the L2 norm of the residuals.}
    \label{fig:corr_spr_l2res_unc_overall}
\end{figure}

Residual is defined to be the difference between the ground truth value and the predicted box parameter value. To further demonstrate the reasonableness of the epistemic uncertainties generated, we calculate their Spearman's rank correlation with the L2 norm of the residual at every epoch during training. In Figure \ref{fig:corr_spr_l2res_unc_overall}, the final epistemic uncertainties of each 3D box parameter is positively correlated with the L2 norm of the residual (correlation with value greater than 25\%), which is the desired result. Among all the 3D parameters, the weakest correlations are exhibited by the box dimensions $l, w$ and $h$, but their decreasing trend is more apparent during training. The strongest positive correlation is exhibited by the rotation (yaw angle).

\subsection{Qualitative Analysis of the Generated Pseudo Labels}

\begin{figure}[!h]
    \centering
    \begin{subfigure}{0.49\textwidth}
        \centering
        \includegraphics[width=7cm]{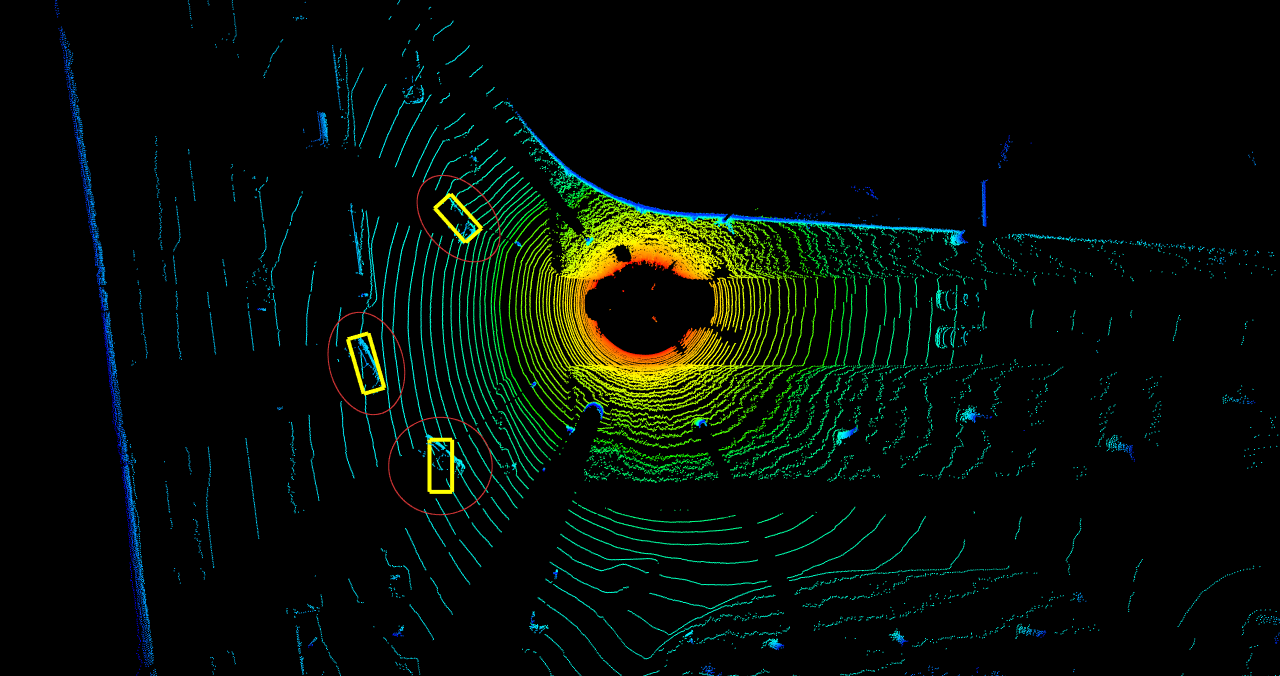}
        \label{fig:medlu_output_1}
    \end{subfigure}
    \hfill
    \begin{subfigure}{0.49\textwidth}
        \centering
        \includegraphics[width=7cm]{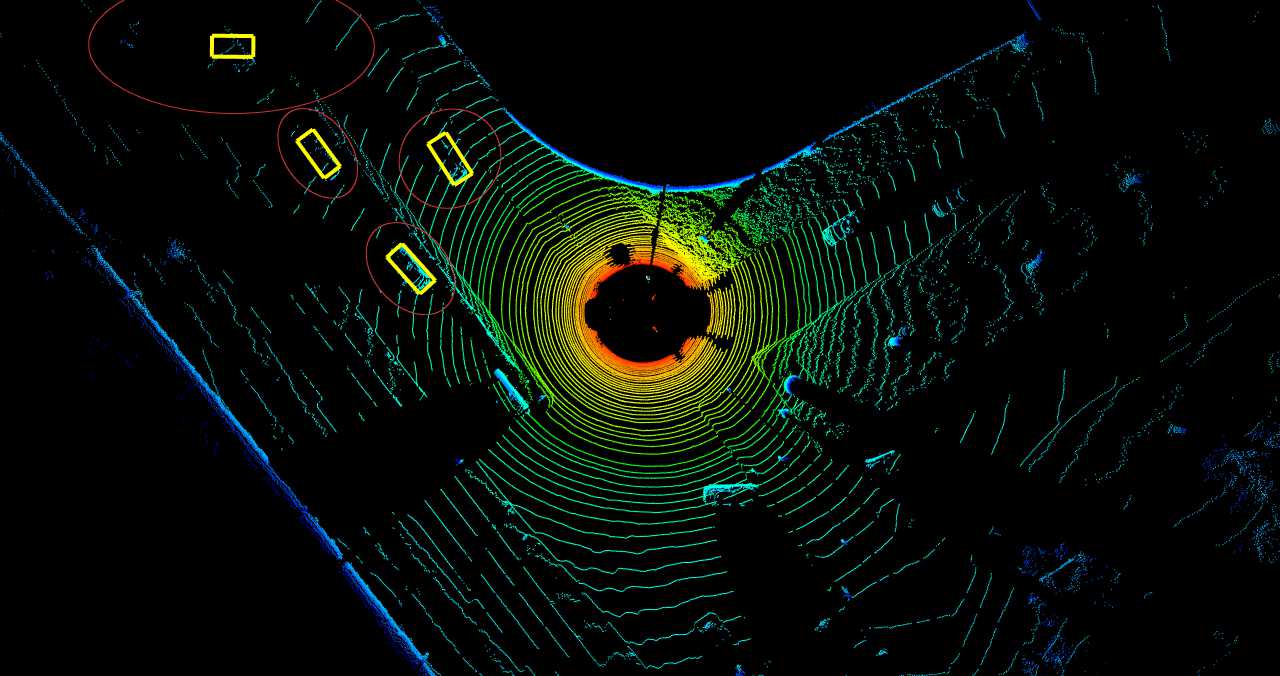}
        \label{fig:medlu_output_2}
    \end{subfigure}
    \hfill
    \begin{subfigure}{0.49\textwidth}
        \centering
        \includegraphics[width=7cm]{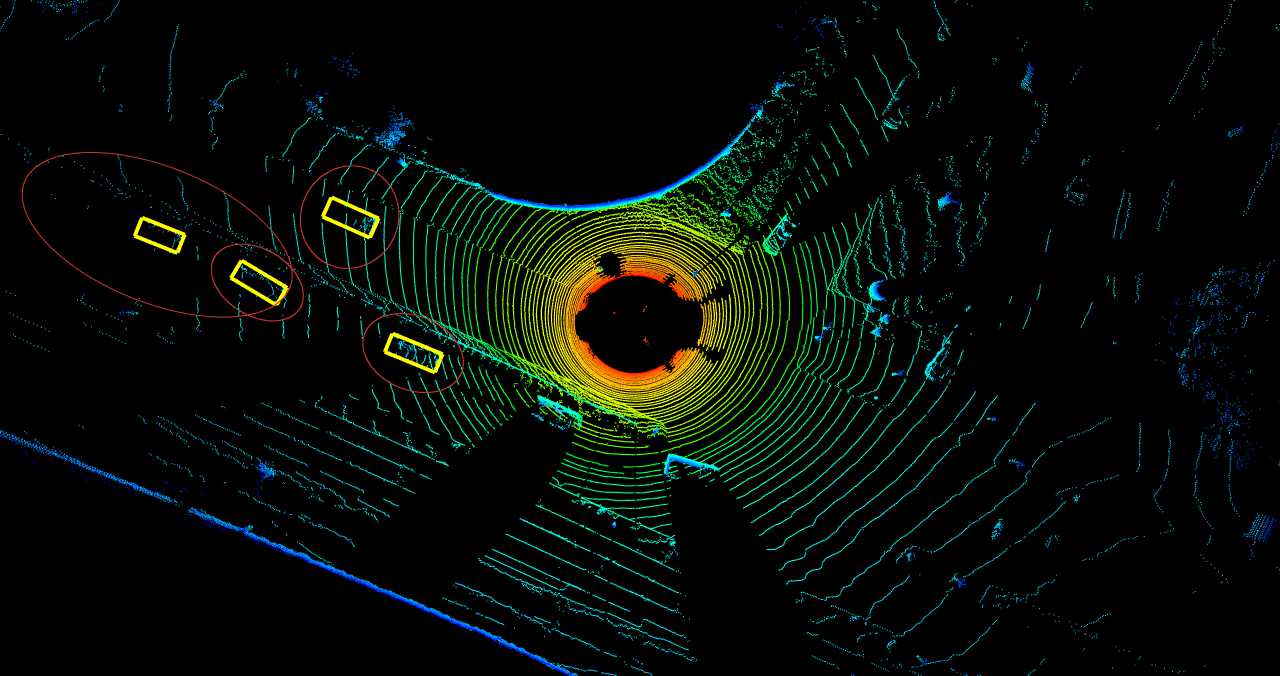}
        \label{fig:medlu_output_3}
    \end{subfigure}
    \caption{Visualization in Birds Eye View (BEV) of the annotation results of MEDL-U on the KITTI tracking data (training set). The red ellipse indicates the magnitude of the uncertainty where the lengths of the axes represent the uncertainty for the length and width of the 3D box.}
    \label{fig:medlu_outputs_bev}
\end{figure}

\begin{figure}[!h]
    \centering
    \begin{subfigure}{0.49\textwidth}
        \centering
        \includegraphics[width=7cm]{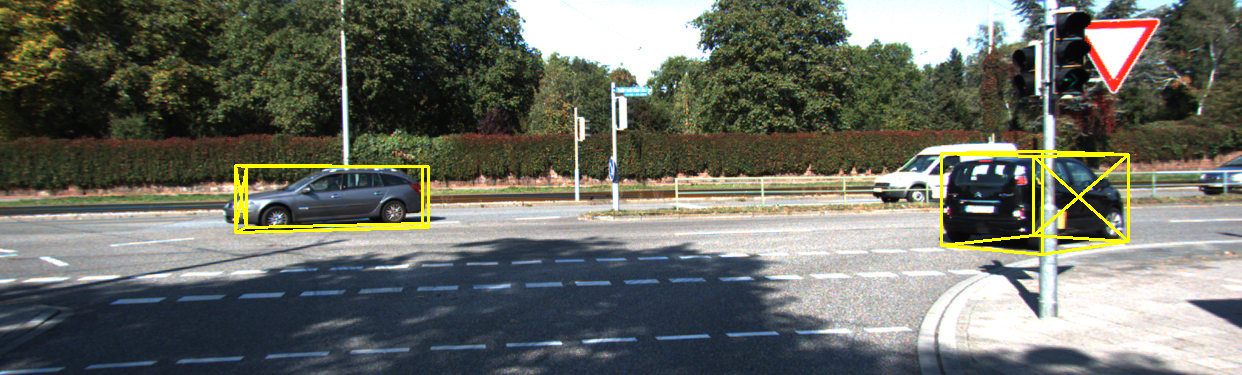}
        \label{fig:medlu_output_1}
    \end{subfigure}
    \hfill
    \begin{subfigure}{0.49\textwidth}
        \centering
        \includegraphics[width=7cm]{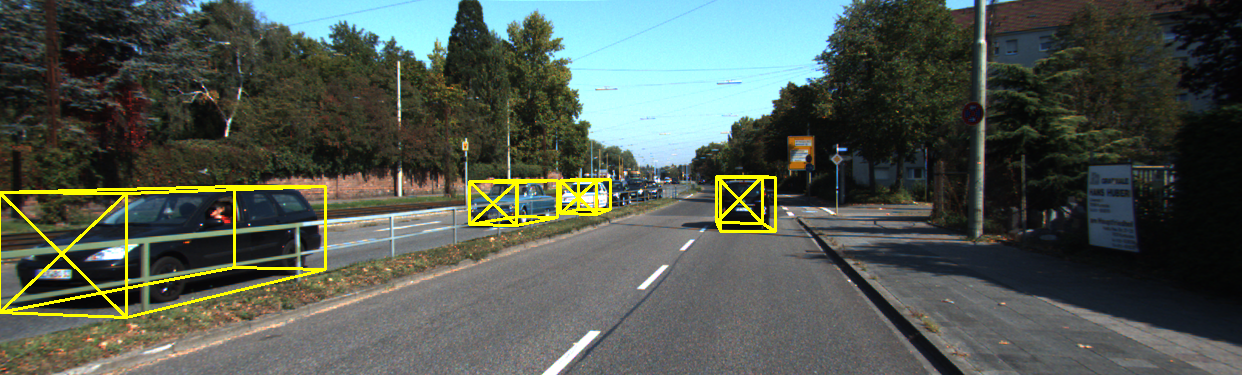}
        \label{fig:medlu_output_2}
    \end{subfigure}
    \hfill
    \begin{subfigure}{0.49\textwidth}
        \centering
        \includegraphics[width=7cm]{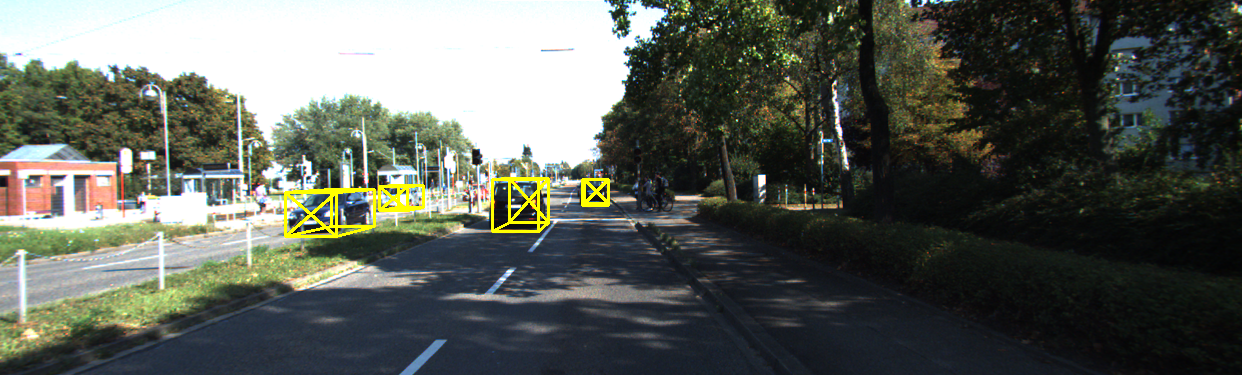}
        \label{fig:medlu_output_3}
    \end{subfigure}
    \caption{Visualization in Camera View of the annotation results of MEDL-U on the KITTI tracking data (training set).}
    \label{fig:medlu_outputs_cam}
\end{figure}

\begin{figure*}[!ht]
    \centering
    
    \begin{subfigure}{0.49\textwidth}
        \centering
        \includegraphics[width=7cm]{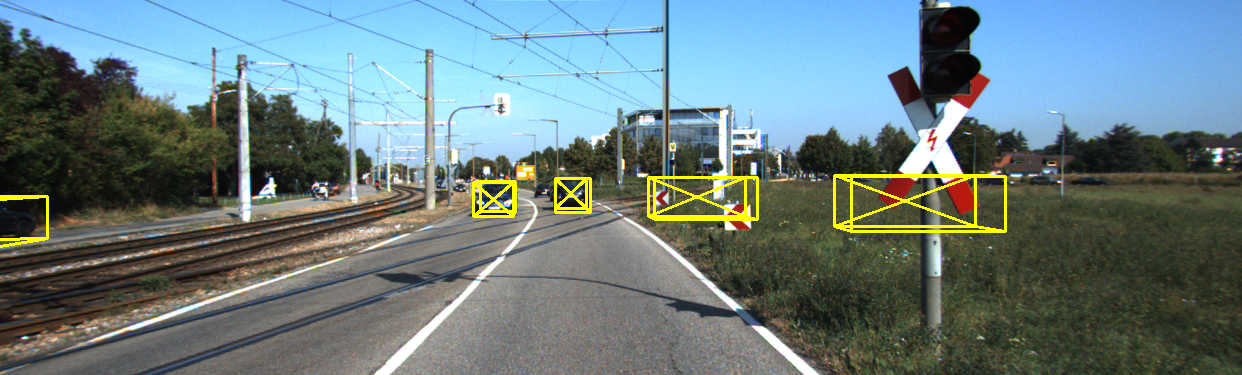}
        \label{fig:medlu_output_1}
        \vspace{-0.3cm}
    \end{subfigure}
    \hfill
    \begin{subfigure}{0.49\textwidth}
        \centering
        \includegraphics[width=7cm]{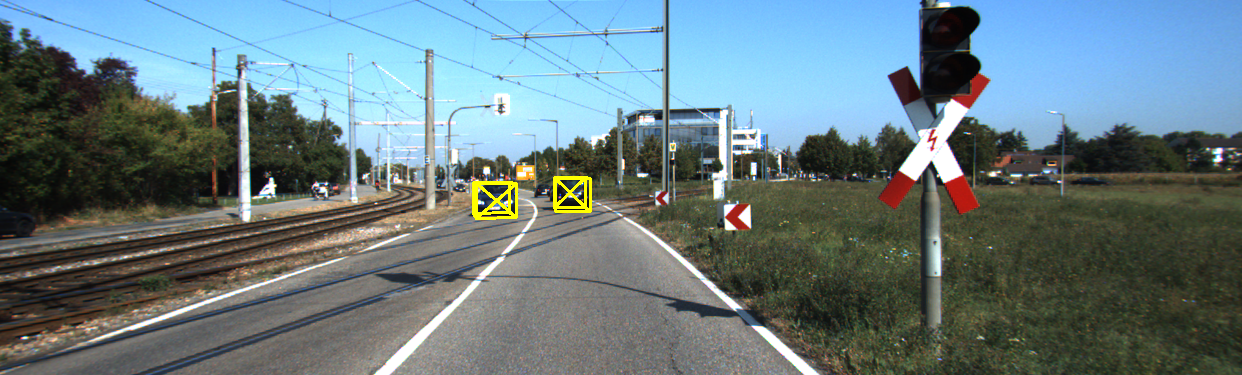}
        \label{fig:medlu_output_2}
        \vspace{-0.3cm}
    \end{subfigure}
    \begin{subfigure}{0.49\textwidth}
        \centering
        \includegraphics[width=7cm]{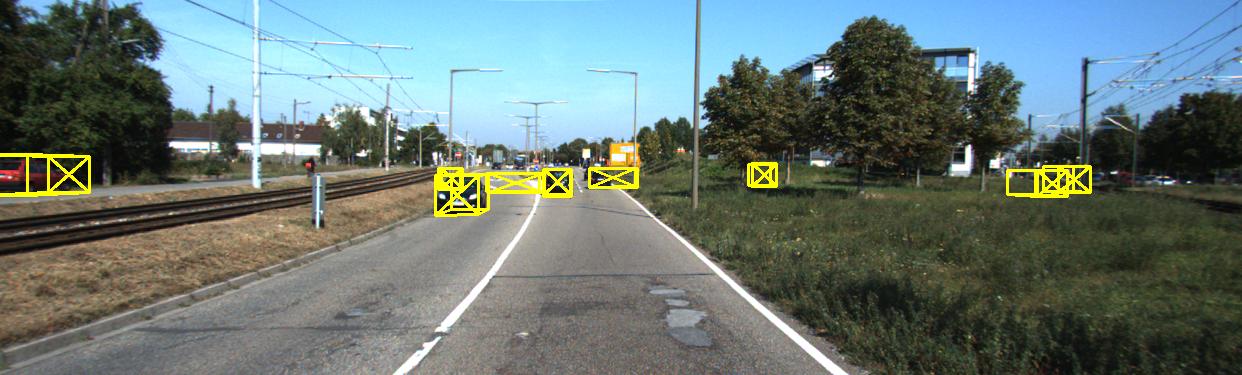}
        \label{fig:medlu_output_3}
        \vspace{-0.3cm}
    \end{subfigure}
    \hfill
    \begin{subfigure}{0.49\textwidth}
        \centering
        \includegraphics[width=7cm]{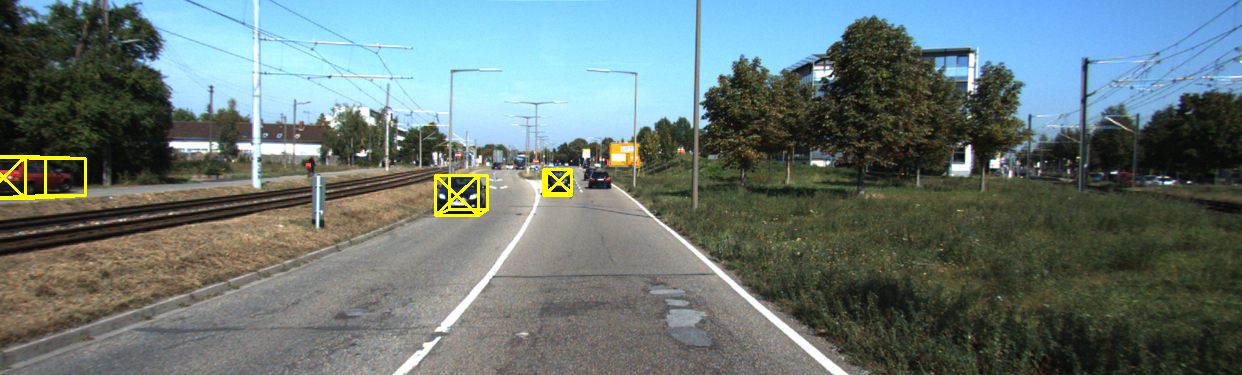}
        \label{fig:medlu_output_3}
        \vspace{-0.3cm}
    \end{subfigure}
    \begin{subfigure}{0.49\textwidth}
        \centering
        \includegraphics[width=7cm]{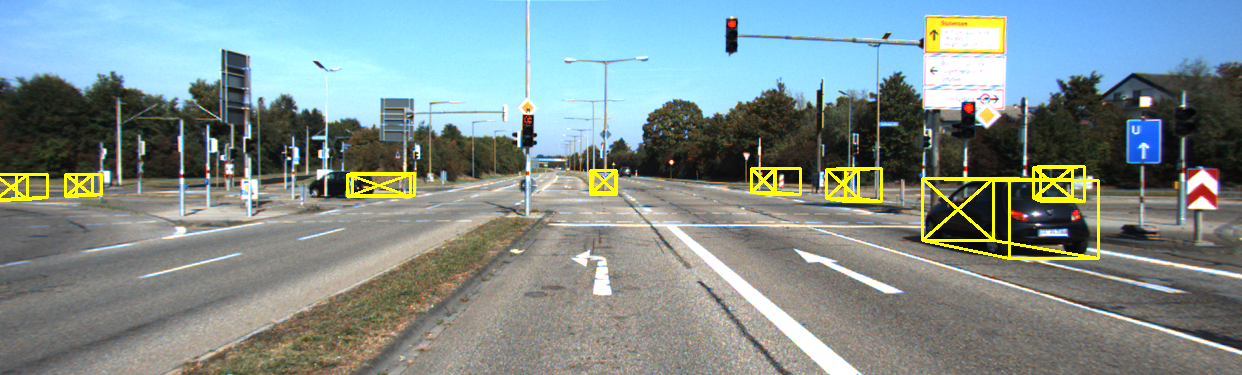}
        \vspace{-0.1cm}
        \caption{trained on MTrans outputs}
        \label{fig:medlu_output_3}
    \end{subfigure}
    \hfill
    \begin{subfigure}{0.49\textwidth}
        \centering
        \includegraphics[width=7cm]{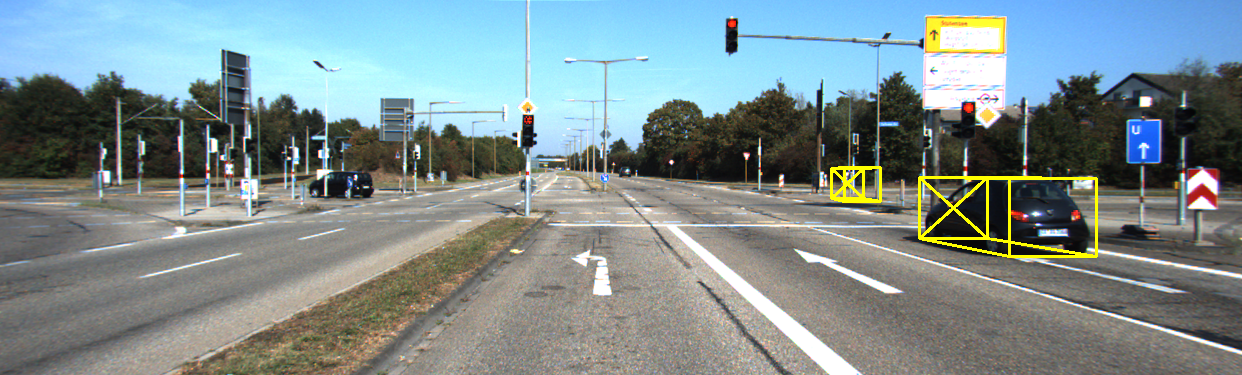}
        \vspace{-0.1cm}
        \caption{trained on MEDL-U outputs}
        \label{fig:medlu_output_3}
    \end{subfigure}
    \caption{Camera visualizations of the detection results of PointPillars when trained using outputs from MTrans (left) and MEDL-U (right). The 3D autolabelers are trained with 500 frames of annotated data.}
    \label{fig:medlu_vs_mtrans_cam}
\end{figure*}

\begin{figure*}[!ht]
    \centering
    \begin{subfigure}{0.49\textwidth}
        \centering
        \includegraphics[width=7cm]{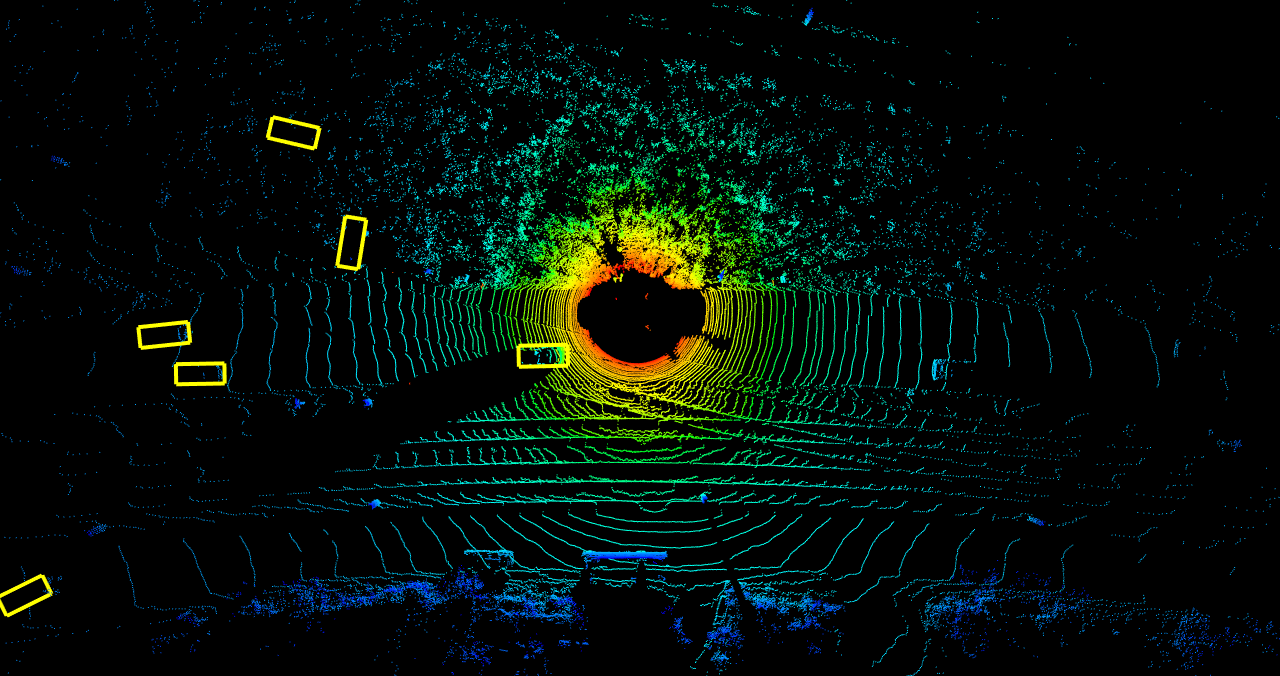}
        \label{fig:medlu_output_1}
        \vspace{-0.3cm}
    \end{subfigure}
    \hfill
    \begin{subfigure}{0.49\textwidth}
        \centering
        \includegraphics[width=7cm]{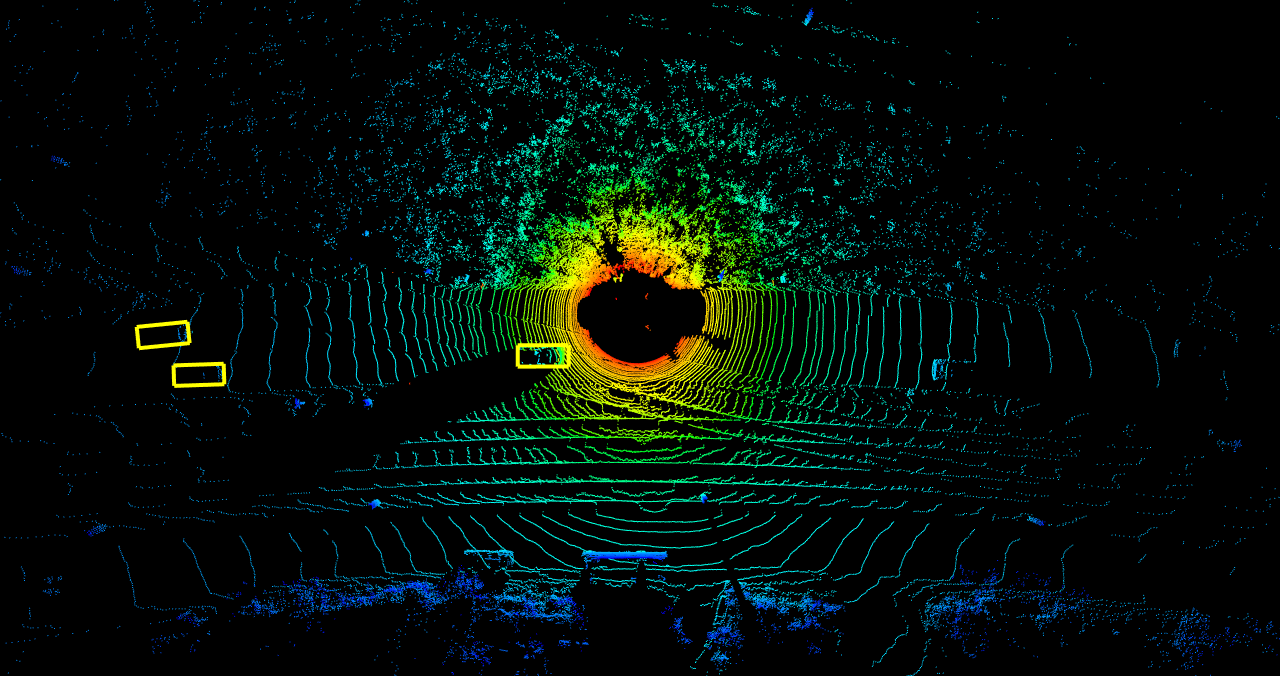}
        \label{fig:medlu_output_2}
        \vspace{-0.3cm}
    \end{subfigure}
    \hfill
    \begin{subfigure}{0.49\textwidth}
        \centering
        \includegraphics[width=7cm]{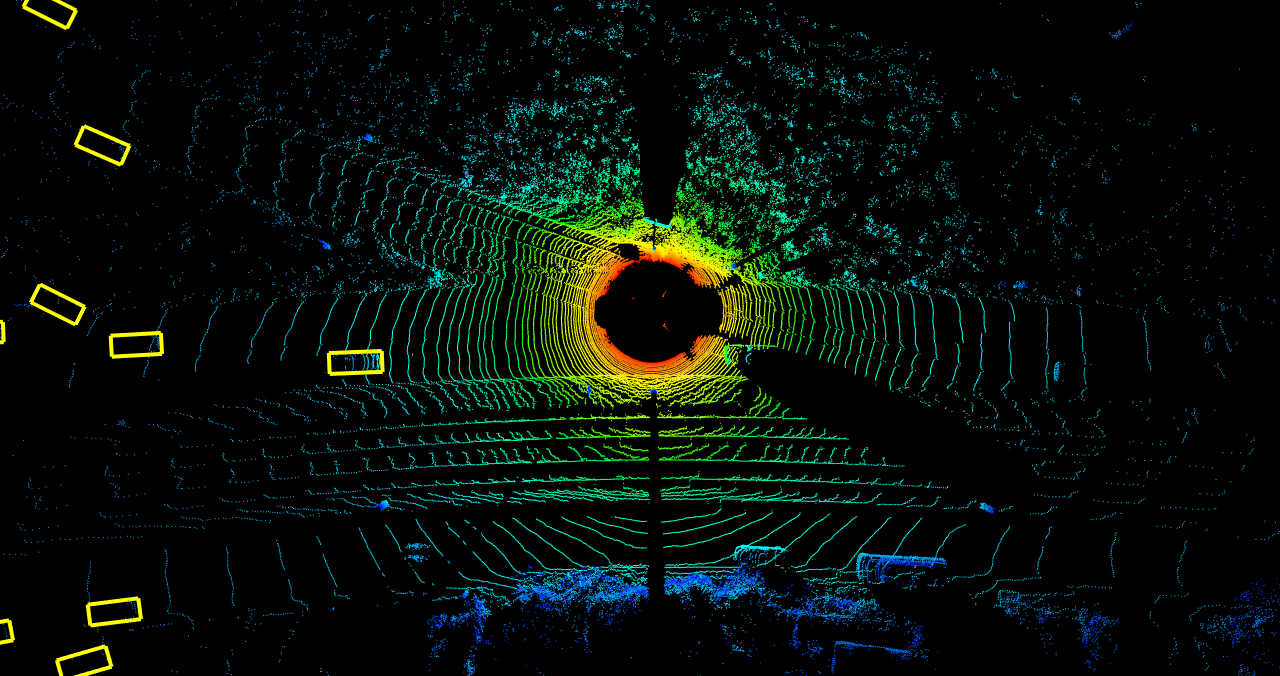}
        \label{fig:medlu_output_3}
        \vspace{-0.3cm}
    \end{subfigure}
    \hfill
    \begin{subfigure}{0.49\textwidth}
        \centering
        \includegraphics[width=7cm]{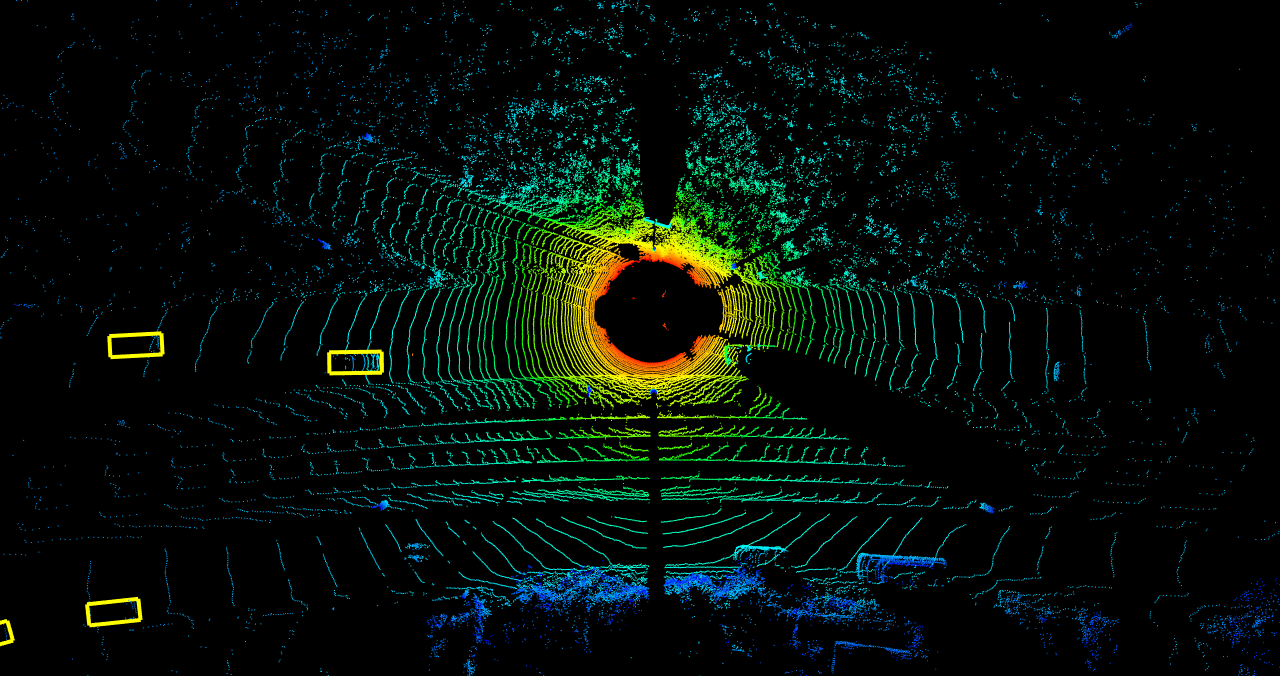}
        \label{fig:medlu_output_3}
        \vspace{-0.3cm}
    \end{subfigure}
    \hfill
    \begin{subfigure}{0.49\textwidth}
        \centering
        \includegraphics[width=7cm]{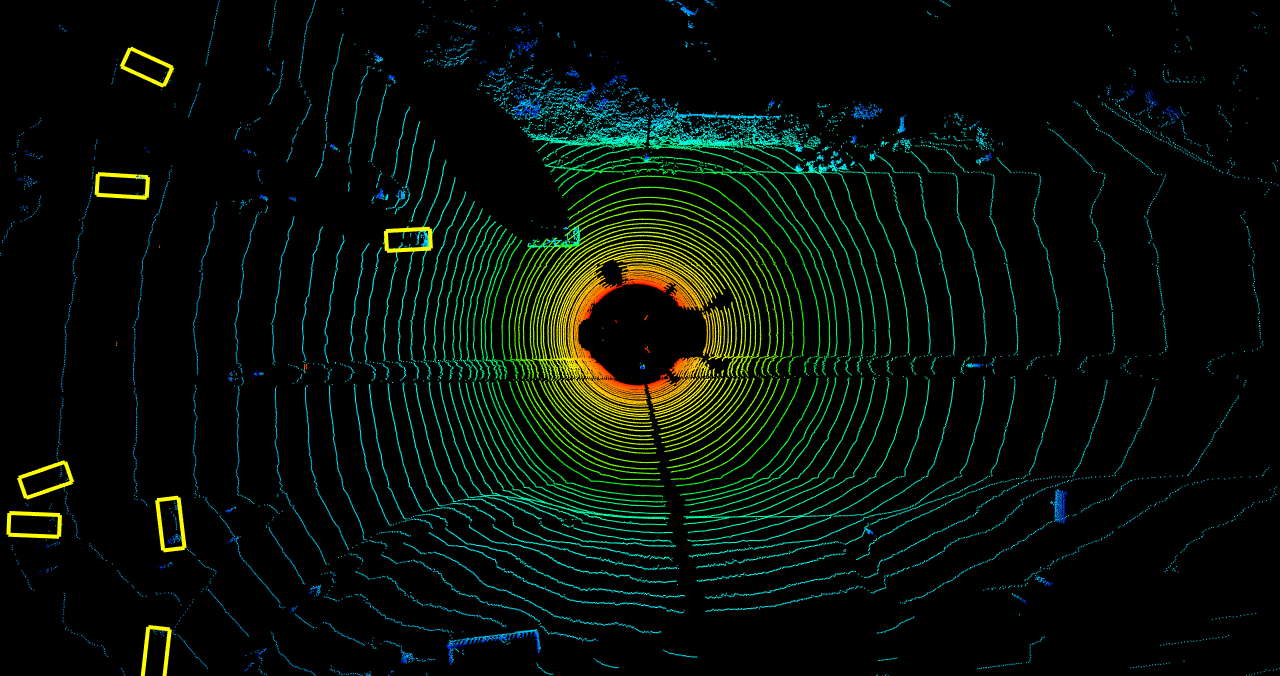}
        \vspace{-0.1cm}
        \caption{trained on MTrans outputs}
        \label{fig:medlu_output_3}
    \end{subfigure}
    \hfill
    \begin{subfigure}{0.49\textwidth}
        \centering
        \includegraphics[width=7cm]{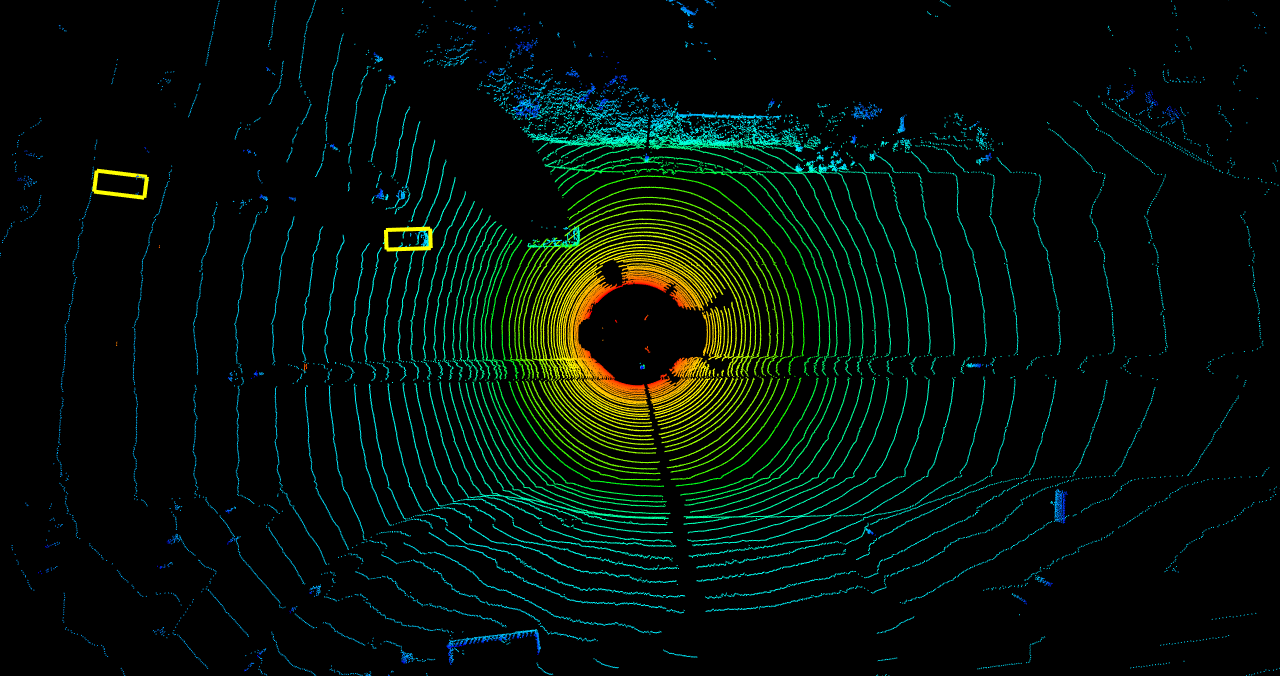}
        \vspace{-0.1cm}
        \caption{trained on MEDL-U outputs}
        \label{fig:medlu_output_3}
    \end{subfigure}
    \caption{BEV visualizations of the detection results of PointPillars when trained using outputs from MTrans (left) and MEDL-U (right). The 3D autolabelers are trained with 500 frames of annotated data.}
    \label{fig:medlu_vs_mtrans_bev}
\end{figure*}

\begin{figure*}[!ht]
    \centering
    \begin{subfigure}{0.49\textwidth}
        \centering
        \includegraphics[width=7cm]{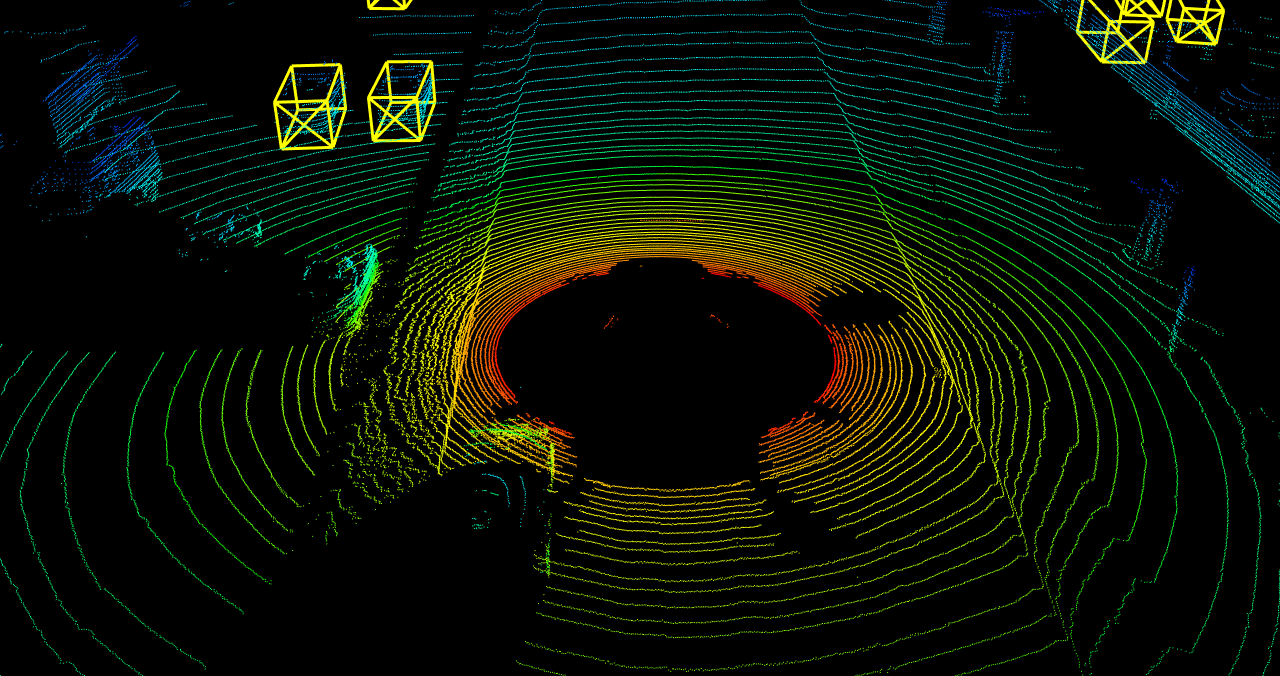}
        \label{fig:medlu_output_1}
        \vspace{-0.3cm}
    \end{subfigure}
    \hfill
    \begin{subfigure}{0.49\textwidth}
        \centering
        \includegraphics[width=7cm]{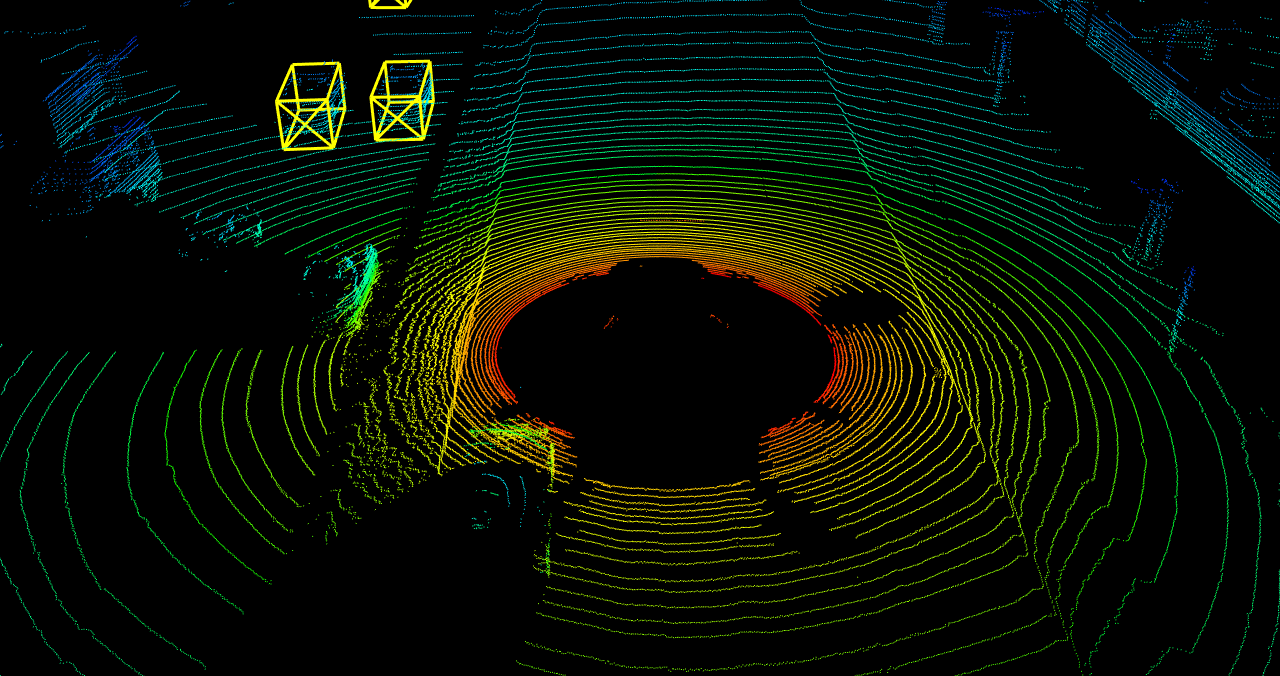}
        \label{fig:medlu_output_2}
        \vspace{-0.3cm}
    \end{subfigure}
    \hfill
    \begin{subfigure}{0.49\textwidth}
        \centering
        \includegraphics[width=7cm]{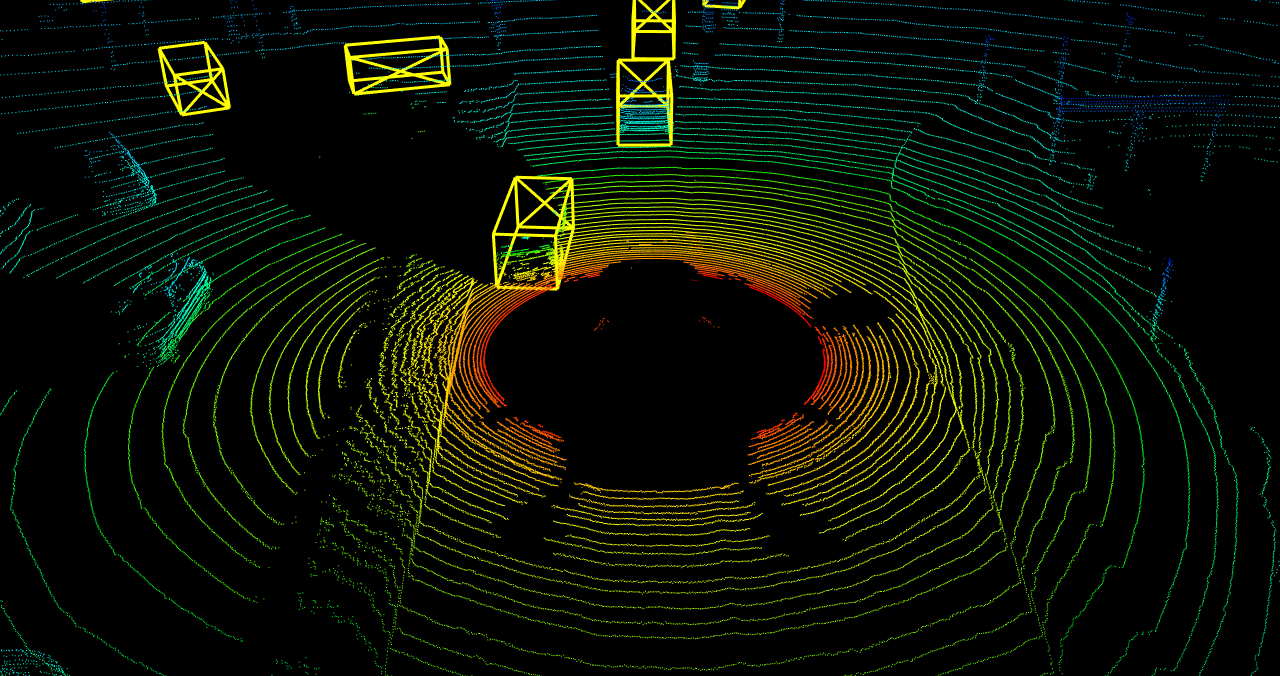}
        \label{fig:medlu_output_3}
        \vspace{-0.3cm}
    \end{subfigure}
    \hfill
    \begin{subfigure}{0.49\textwidth}
        \centering
        \includegraphics[width=7cm]{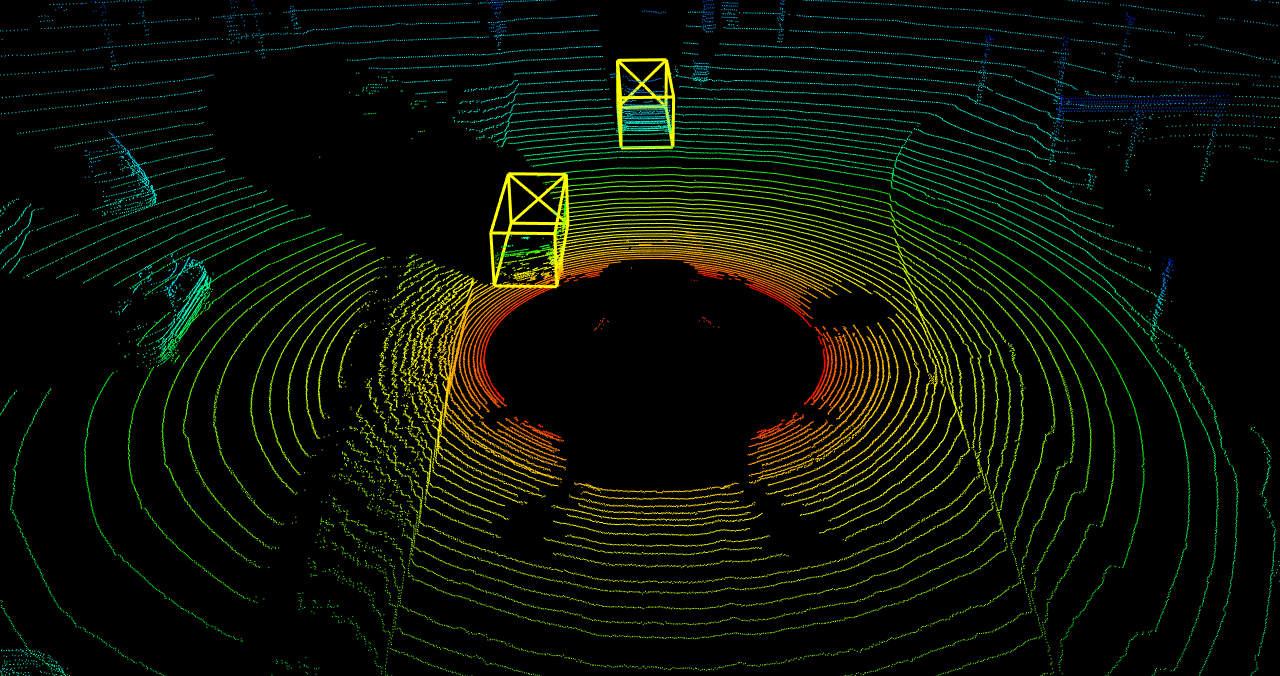}
        \label{fig:medlu_output_3}
        \vspace{-0.3cm}
    \end{subfigure}
    \hfill
    \begin{subfigure}{0.49\textwidth}
        \centering
        \includegraphics[width=7cm]{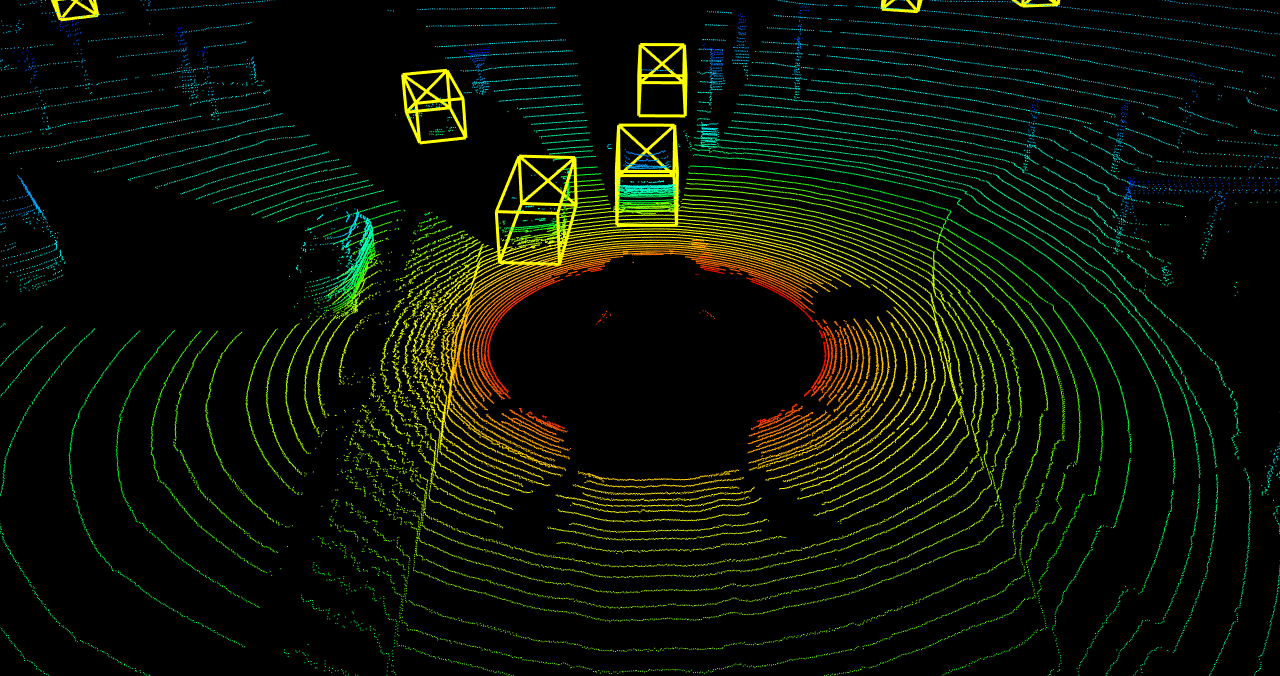}
        \vspace{-0.1cm}
        \caption{trained on MTrans outputs}
        \label{fig:medlu_output_3}
    \end{subfigure}
    \hfill
    \begin{subfigure}{0.49\textwidth}
        \centering
        \includegraphics[width=7cm]{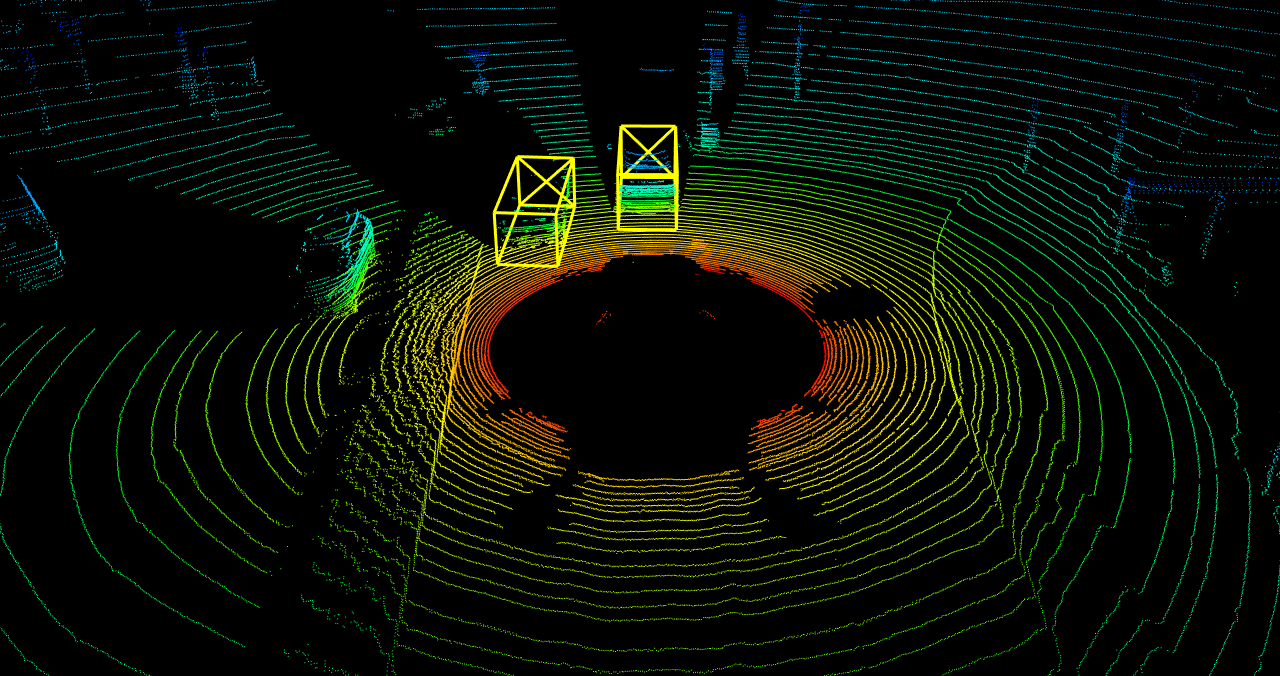}
        \vspace{-0.1cm}
        \caption{trained on MEDL-U outputs}
        \label{fig:medlu_output_3}
    \end{subfigure}
    \caption{Point cloud visualizations of the detection results of PointPillars when trained using outputs from MTrans (left) and MEDL-U (right). The 3D autolabelers are trained with 125 frames of annotated data.}
    \label{fig:medlu_vs_mtrans_pc125}
\end{figure*}

\begin{figure*}[!ht]
    \centering
    \begin{subfigure}{0.49\textwidth}
        \centering
        \includegraphics[width=7cm]{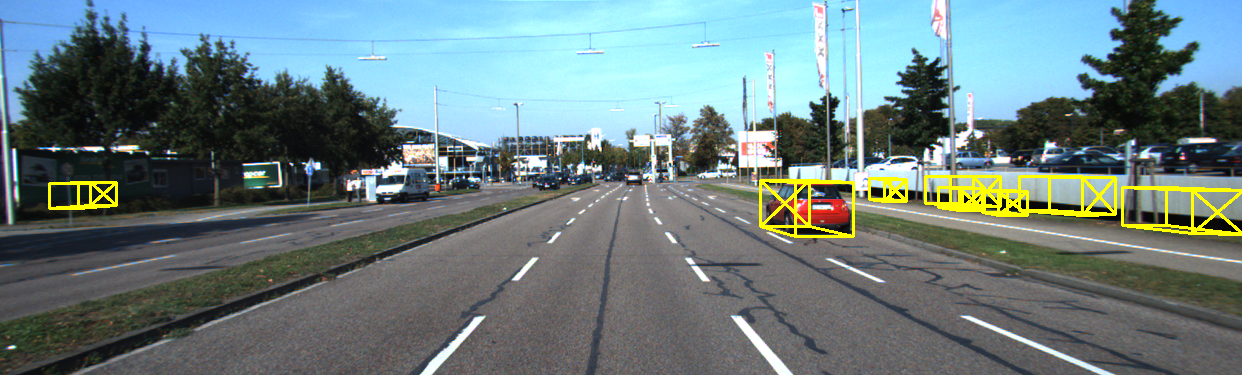}
        \label{fig:medlu_output_1}
        \vspace{-0.3cm}
    \end{subfigure}
    \hfill
    \begin{subfigure}{0.49\textwidth}
        \centering
        \includegraphics[width=7cm]{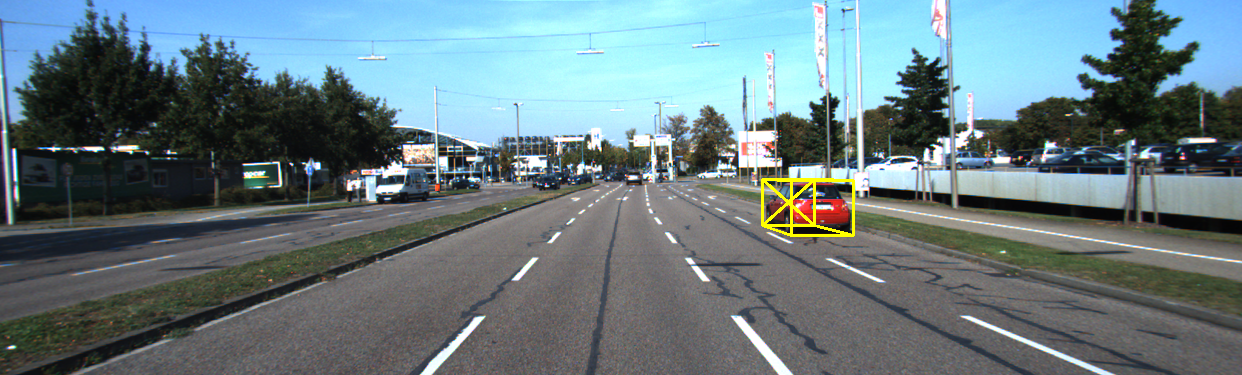}
        \label{fig:medlu_output_2}
        \vspace{-0.3cm}
    \end{subfigure}
    \hfill
    \begin{subfigure}{0.49\textwidth}
        \centering
        \includegraphics[width=7cm]{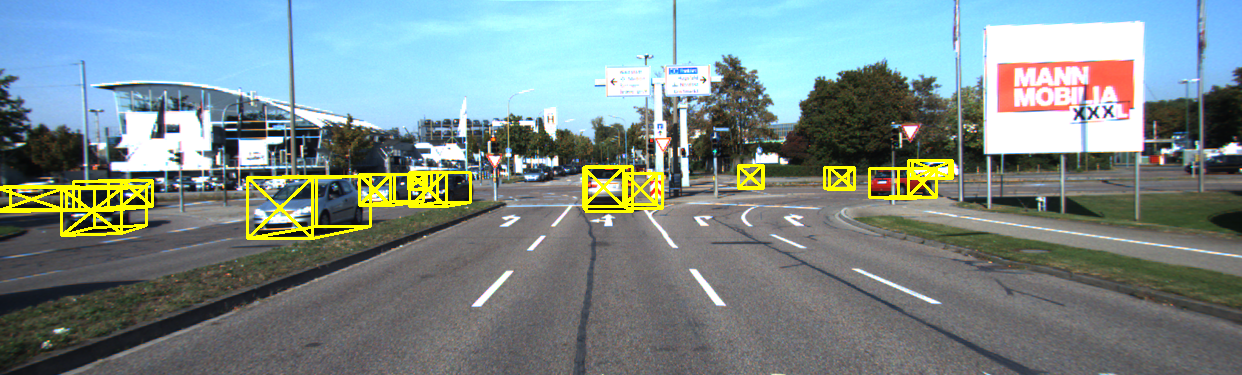}
        \label{fig:medlu_output_3}
        \vspace{-0.3cm}
    \end{subfigure}
    \hfill
    \begin{subfigure}{0.49\textwidth}
        \centering
        \includegraphics[width=7cm]{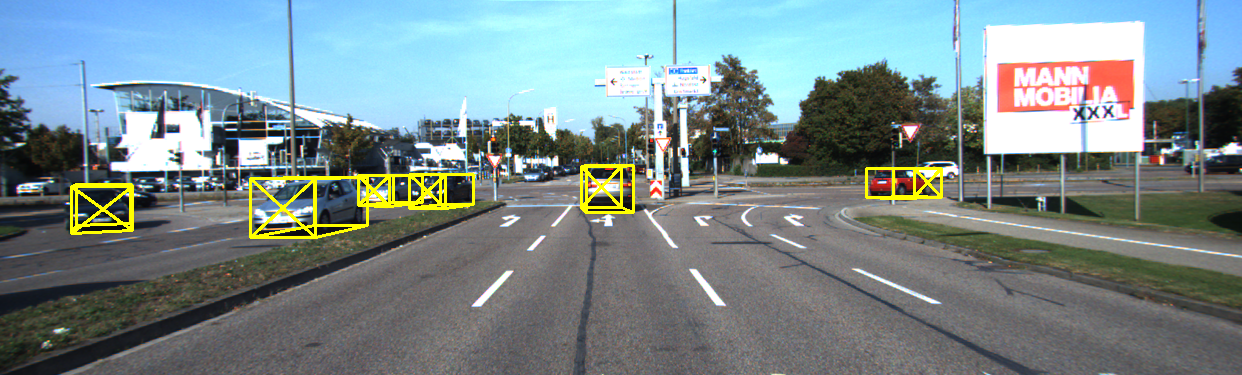}
        \label{fig:medlu_output_3}
        \vspace{-0.3cm}
    \end{subfigure}
    \hfill
    \begin{subfigure}{0.49\textwidth}
        \centering
        \includegraphics[width=7cm]{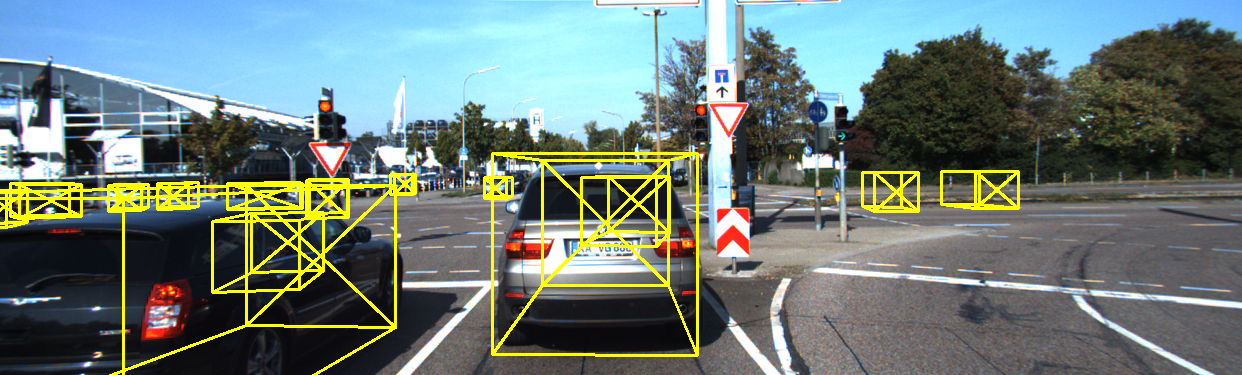}
        \vspace{-0.1cm}
        \caption{trained on MTrans outputs}
        \label{fig:medlu_output_3}
    \end{subfigure}
    \hfill
    \begin{subfigure}{0.49\textwidth}
        \centering
        \includegraphics[width=7cm]{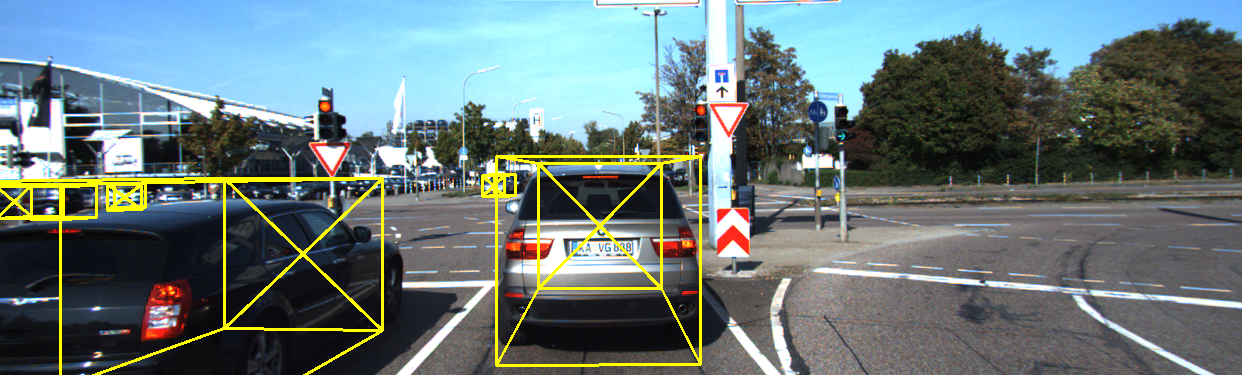}
        \vspace{-0.1cm}
        \caption{trained on MEDL-U outputs}
        \label{fig:medlu_output_3}
    \end{subfigure}
    \caption{Camera visualizations of the detection results of PointPillars when trained using outputs from MTrans (left) and MEDL-U (right). The 3D autolabelers are trained with 125 frames of annotated data.}
    \label{fig:medlu_vs_mtrans_cam125}
\end{figure*}

In this section, we visualize the annotation results of MEDL-U and detection results of the 3D detector PointPillars trained with the outputs of MTrans and MEDL-U. In Figures \ref{fig:medlu_vs_mtrans_cam} and \ref{fig:medlu_vs_mtrans_bev}, we provide visual comparison of the detection results of PointPillars trained with MTrans pseudo labels and MEDL-U pseudo labels and uncertainties on the KITTI tracking data (test set) when the 3D autolabelers are trained with 500 frames of annotated data. Visual comparisons are also shown in Figures \ref{fig:medlu_vs_mtrans_pc125} and \ref{fig:medlu_vs_mtrans_cam125} for the case when the autolabelers are trained with 125 frames of annotated data. It can be seen that PointPillars trained with MEDL-U obtains better detection results. In Figures \ref{fig:medlu_outputs_bev} and \ref{fig:medlu_outputs_cam}, the annotation results of MEDL-U on the KITTI tracking data (training set) are visualized. Visualizing in BEV, we observe that the proposed framework generates reasonable uncertainty estimates associated with each object without compromising the quality of the pseudo labels. We can see that in most cases, MEDL-U generates larger uncertainty estimates for faraway objects with sparser points. 

\clearpage